\documentclass{article}
\usepackage{stair_preprint,times}

\usepackage{amsmath,amsfonts,bm}

\def\eqref#1{equation~\ref{#1}}

\def\1{\bm{1}}

\DeclareMathAlphabet{\mathsfit}{\encodingdefault}{\sfdefault}{m}{sl}
\SetMathAlphabet{\mathsfit}{bold}{\encodingdefault}{\sfdefault}{bx}{n}

\usepackage{graphicx}
\usepackage{wrapfig}
\usepackage{booktabs}
\usepackage{float}
\usepackage{xcolor}
\usepackage{colortbl}
\definecolor{stairrow}{RGB}{235,244,240}
\definecolor{vanillarow}{gray}{0.9}
\newcommand{\meanstd}[2]{\mbox{#1\textsubscript{\normalfont\ensuremath{\pm#2}}}}
\usepackage{amssymb}
\usepackage{hyperref}
\hypersetup{
  hidelinks,
  pdftitle={Can Computation from Earlier Problems Help LLMs Solve New Ones?},
  pdfauthor={Jipei He, Wenhui Tan, Xiaoyi Yu, Enver Sangineto, Fiorenzo Parascandolo, Rita Cucchiara, Ruihua Song}
}
\usepackage{url}

\title{Can Computation from Earlier Problems Help LLMs Solve New Ones?}
\author{Jipei He\textsuperscript{1}\quad Wenhui Tan\textsuperscript{1}\quad
Xiaoyi Yu\textsuperscript{1}\quad Enver Sangineto\textsuperscript{2}\\
Fiorenzo Parascandolo\textsuperscript{2}\quad Rita Cucchiara\textsuperscript{2}\quad
Ruihua Song\textsuperscript{1,*}\\[0.4em]
\textsuperscript{1}Gaoling School of Artificial Intelligence, Renmin University of China, Beijing, China\\
\textsuperscript{2}University of Modena and Reggio Emilia, Italy\\
\textsuperscript{*}Corresponding author: \texttt{rsong@ruc.edu.cn}}

\begin{document}
\maketitle
\begin{abstract}
Large language models often solve independent problems in the same conversation. Can computation from earlier problems help them solve new ones? Preliminary experiments show that retained history can raise or lower later-turn accuracy, even within the same domain. To trace this effect, we hold each problem fixed and vary its preceding history. The resulting changes in the model's internal states contain a recurring, problem-dependent component. To improve reasoning with retained history, we introduce \textbf{STAIR} (\textbf{Stale-Token Attention for Inter-query Reuse}). It captures attention keys and values from earlier response generation in a read-only bank. STAIR learns to reflect current query vectors to redirect attention over this bank during prompt processing. The backbone and stored entries remain fixed; only 12,288 parameters are trained. Across three Qwen models and four reasoning benchmarks, STAIR improves average later-turn accuracy by up to 11.67 percentage points over the unmodified model with history. The math-trained Qwen3.5-9B controller also improves code generation on HumanEval+ without further training.
\end{abstract}

\section{Introduction}\label{sec:1}

Users often move between unrelated questions within the same conversation with a large language model (LLM).
Earlier exchanges remain visible as text, so users can return to previous discussions after switching tasks.
For the model, answering those earlier questions also involved a sequence of internal computations.
Processing the retained conversation makes earlier token positions available to current queries through attention. 
Language models already reuse representations from earlier inputs \citep{dai2019transformerxl,wu2022memorizing}.
Whether earlier computation remains useful after a task switch is less clear.
In this paper, we ask:
\textbf{\textit{Can the historical computation left by an earlier problem still help solve the current one?}}

\begin{figure}[t]
    \centering
    \includegraphics[width=\linewidth]{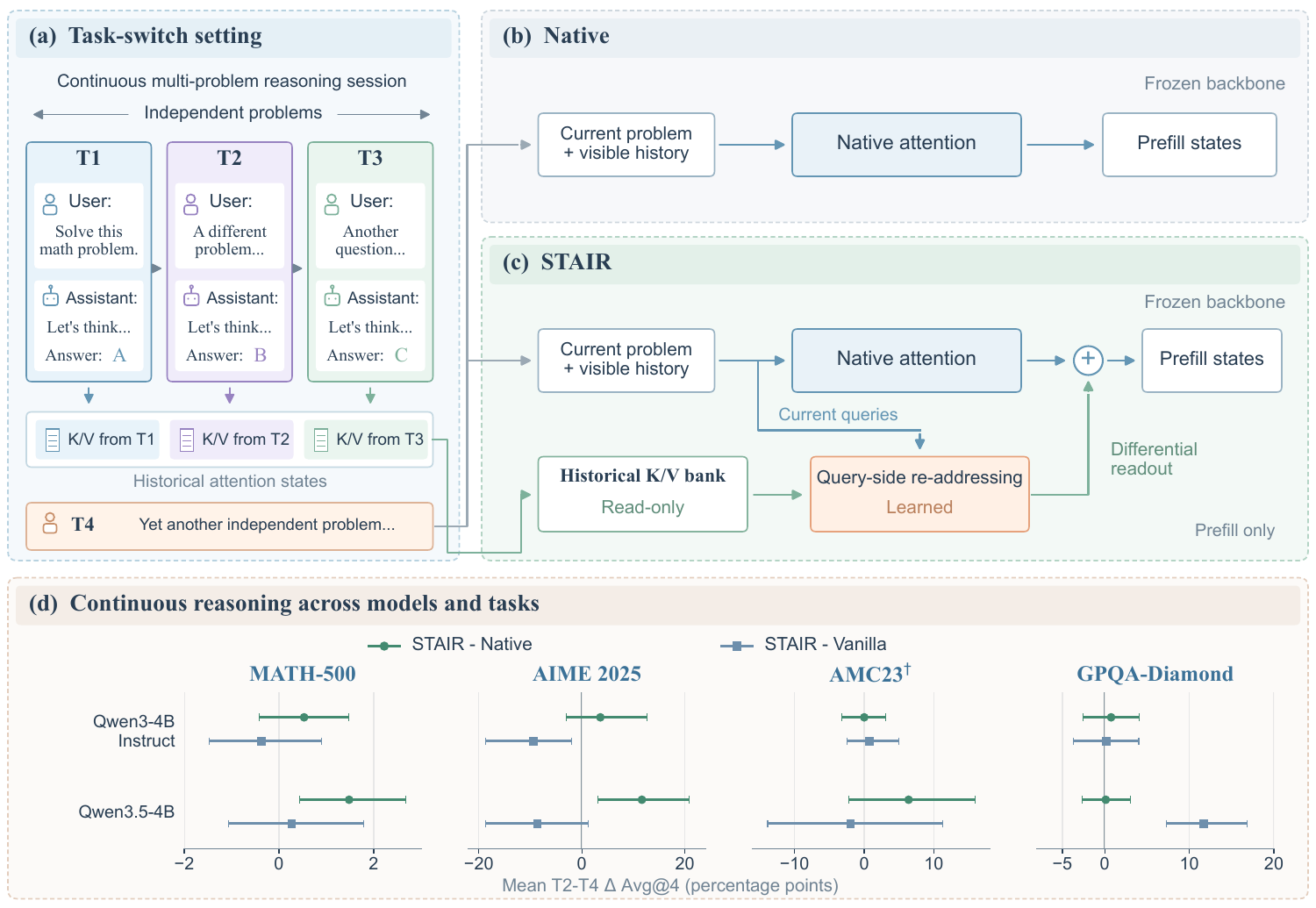}
    \caption{STAIR in a continuous problem-solving session. \textbf{(a)} Earlier problems leave conversation history; the assistant responses illustrate Instruct, where reasoning and answers remain visible. T4 introduces a new independent problem. \textbf{(b)} Native is the unmodified model reading that history. \textbf{(c)} STAIR changes how current queries read a separate bank of attention keys and values captured during earlier turns. It acts during prompt processing (prefill), with model weights fixed. Both conditions process the visible conversation anew each turn. \textbf{(d)} Paired differences in mean T2--T4 Avg@4 against Native and the no-history Vanilla reference, with pointwise 95\% intervals (Appendix~\ref{app:paired-intervals}). Positive values favor STAIR; horizontal scales vary by task. $\dagger$ AMC23 excludes training-overlap problems, leaving 34 problems.}
    \label{fig:overview}
\end{figure}

We place the same problem set at different positions in four-turn sessions, with the sampling configuration and checkpoint fixed. Each new problem is independent of the earlier ones and supplies its own task-specific information. Vanilla answers without history; Native is the unmodified model with earlier turns in context. Avg@4 averages correctness over four sampled responses per problem. History lowers accuracy on MATH-500 \citep{lightman2023verify} with Qwen3-4B Instruct \citep{qwen2025instruct2507}, yet raises it on GPQA-Diamond \citep{rein2024gpqa} with Qwen3.5-4B \citep{qwen2026qwen35}.

To trace these effects, we replay histories with Qwen3-4B Instruct and the hybrid-attention Qwen3.5-4B. Current queries allocate attention to earlier assistant responses, and history shifts the current problem's representation. After we remove separate problem and history effects, many of the same problems remain nearest neighbors across histories. At the response level, retained history can also change answers in both directions, correcting some errors while introducing others.

\begin{samepage}
To learn how to use earlier assistant states, we introduce \textbf{STAIR}, short for \textbf{S}tale-\textbf{T}oken \textbf{A}ttention for \textbf{I}nter-query \textbf{R}euse (Figure~\ref{fig:overview}).
Stale tokens are assistant tokens from completed problems. STAIR stores their attention keys and values (K/V) in a read-only bank. For thinking models, the bank retains reasoning states while the conversation retains the answer text. Queries are vectors that weight the keys to read their values. STAIR reflects each query by reversing its component along a learned direction, changing these weights. The backbone and stored entries stay fixed. STAIR acts only during prompt processing (prefill); subsequent generation follows the original decoding path.

\end{samepage}

To evaluate this learned readout, we compare STAIR with Native in four-turn sessions. With 12,288 trainable parameters, STAIR improves Qwen3.5-4B's mean T2--T4 Avg@4 on AIME 2025 \citep{zhang2025aime} by 11.67 percentage points.

Our contributions are:
\begin{enumerate}
    \item We find a recurring, problem-dependent component in history-conditioned representation changes across distinct source histories, and reproduce it in full-attention and hybrid-attention backbones.

    \item We introduce STAIR, which learns query reflections to change how the current problem reads a fixed historical K/V bank while keeping the backbone frozen.

    \item Across three Qwen models and four reasoning benchmarks, STAIR improves mean T2--T4 Avg@4 over Native by up to 11.67 percentage points with only 12,288 trainable parameters.
\end{enumerate}

\clearpage
\section{Historical Computation After Task Switches}\label{sec:historical-computation}

\begin{figure}[H]
    \centering
    \includegraphics[width=\linewidth]{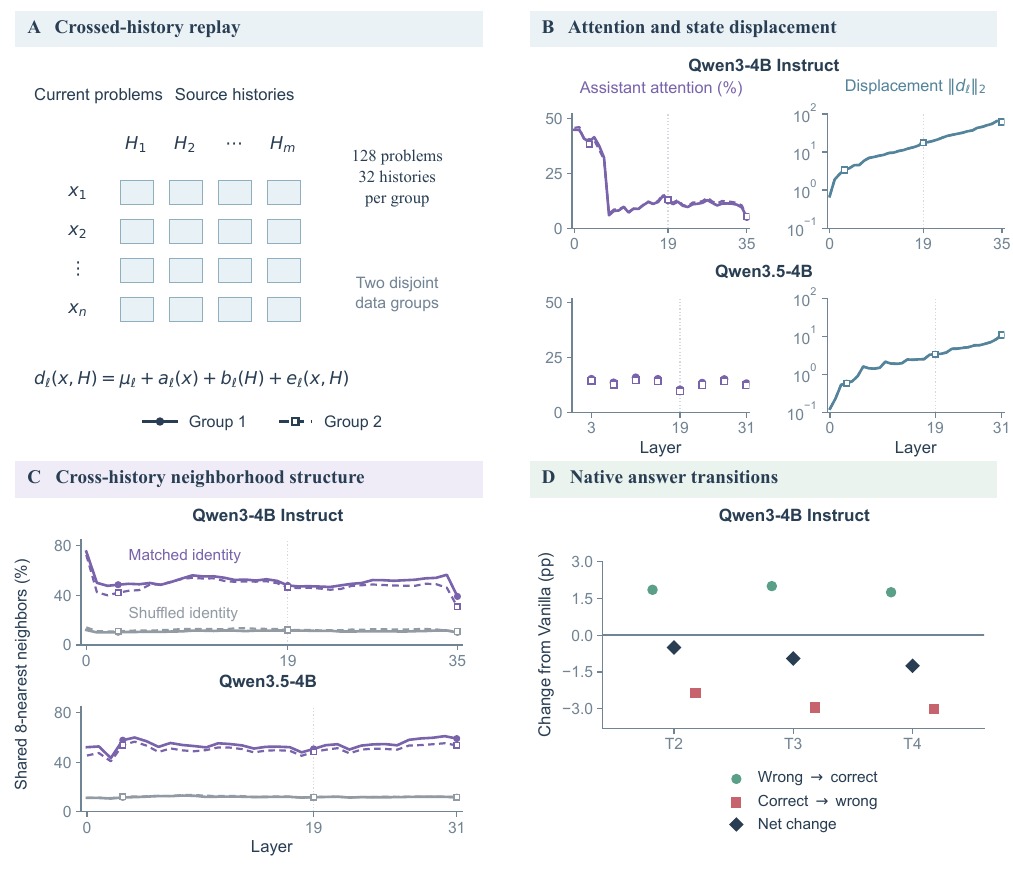}
    \caption{Historical computation after task switches on MATH-500 with Qwen3-4B Instruct and Qwen3.5-4B.
    \textbf{A:} Crossed-history replay with a frozen backbone. Each of two disjoint data groups crosses 128 current problems with 32 three-turn source histories. State displacement is decomposed into additive effects and an interaction response.
    \textbf{B:} Attention mass on visible earlier assistant-response bodies (left) and state-displacement norm (right; logarithmic scale), with one row per model. Qwen3.5 attention points cover its eight full-attention layers.
    \textbf{C:} Cross-history overlap of eight nearest candidate problems in interaction-response space, with the Question-identity shuffle control. At layer 19, interaction energy fractions for groups 1 and 2 are 0.69\% and 0.66\% in Qwen3, and 1.39\% and 1.57\% in Qwen3.5. In B and C, groups 1 and 2 use filled circles and hollow squares, with solid and dashed lines where connected. Vertical dotted lines mark layer 19; indices are zero-based.
    \textbf{D:} Qwen3-4B Instruct Native answer transitions relative to Vanilla T1, normalized by 2,000 pairs per turn. Wrong-to-correct contributions are positive and correct-to-wrong contributions negative. Diamonds mark net changes.}
    \label{fig:historical-computation}
\end{figure}

We hold the current problem fixed and vary the preceding conversation, which we call the \textit{source history}. Each frozen model replays its own saved three-turn histories before processing the current MATH-500 problem. The current problem never appears in its source history. Two data groups use disjoint current problems and history sources (Figure~\ref{fig:historical-computation}A). We study both Qwen3-4B Instruct and Qwen3.5-4B to compare full-attention and hybrid architectures. Appendix~\ref{app:historical-computation} gives the measurement details.

\begin{samepage}
\subsection{Attention to Earlier Responses and State Displacement}\label{sec:historical-reading}

To measure attention allocation after a task switch, we sum the attention weights that queries from the current user-message body assign to earlier assistant-response bodies. We call this share of attention over the full causal context \textit{attention mass}. Instruct retains complete responses, including reasoning; Qwen3.5 retains the text after the thinking span. We also measure the change in the current problem's representation. Let $h_\ell(x\mid H)$ be block $\ell$'s output, averaged over the current problem's body tokens. Its \textit{state displacement} is
\begin{equation}
d_\ell(x,H)=h_\ell(x\mid H)-h_\ell(x\mid\varnothing).
\label{eq:state-displacement}
\end{equation}
The no-history condition retains the same system prompt and current-turn template.
\end{samepage}

Attention remains on earlier responses through middle and later layers (Figure~\ref{fig:historical-computation}B). Instruct assigns about 39\% to them at layer 3 and 13\% at layer 19, with similar profiles in both data groups. Qwen3.5-4B assigns 9.68--15.94\% across its eight full-attention layers. State displacement persists into later layers in both models.

\subsection{Interaction Responses Retain Problem-Dependent Structure}\label{sec:interaction-structure}

Do similar problems respond similarly to different histories? A shift shared by every problem could preserve their neighborhoods simply by moving them together. We therefore isolate the change specific to each problem--history pairing:
\begin{equation}
d_\ell(x,H)=\mu_\ell+a_\ell(x)+b_\ell(H)+e_\ell(x,H),
\label{eq:interaction-decomposition}
\end{equation}
Here $\mu_\ell$ is the overall mean, and $a_\ell(x)$ and $b_\ell(H)$ capture the separate problem and history effects. The remainder $e_\ell(x,H)$ is the \textit{interaction response}. At layer 19 of Instruct, it accounts for 0.66--0.69\% of displacement energy across the two groups, measured by sums of squared norms.

We find each problem's eight nearest neighbors by Euclidean distance between interaction responses, then repeat under a history built from different earlier problems. Neighborhood overlap is the fraction of neighbors shared across the two histories. The control pairs neighborhoods anchored by shuffled problem identities.

At layer 19, overlap reaches 46.50--48.06\% for Instruct, versus about 12\% after shuffling (Figure~\ref{fig:historical-computation}C). Qwen3.5-4B shows 48.22--50.66\% overlap at the same layer, where the interaction accounts for 1.39--1.57\% of energy. Its overlap exceeds the identity-shuffle control throughout all 32 layers (Appendix~\ref{app:qwen35-replication}). A small component of the change thus preserves recurring relationships among current problems across different histories.

\subsection{Native Continuous Reasoning Changes Answers in Both Directions}\label{sec:native-accuracy}

\begin{table}[htbp]
\centering
\setlength{\tabcolsep}{7pt}
\caption{Later-turn accuracy (\%) grouped by the actual first answer in each session. Responses are pooled over T2--T4, three Qwen models, and three mathematical benchmarks; $n$ counts responses. $\Delta$ is STAIR $-$ Native in percentage points.}
\label{tab:history-correctness-summary}
\begin{tabular}{@{}lccccc@{}}
\toprule
Historical T1 & $n$ & Vanilla & Native & STAIR & $\Delta$ (pp)\\
\midrule
Correct & 18,879 & 93.96 & 93.42 & \cellcolor{stairrow}\textbf{94.18} & +0.76\\
Wrong & 1,425 & 80.07 & 60.35 & \cellcolor{stairrow}\textbf{68.56} & +8.21\\
\bottomrule
\end{tabular}
\end{table}
We pair each later Native response with its Vanilla answer to the same problem and sample. Qwen3-4B Instruct on MATH-500 yields 2,000 pairs per turn. At T4, 35 answers change from wrong to correct and 60 from correct to wrong, a net loss of 1.25 percentage points (Figure~\ref{fig:historical-computation}D).

Grouping sessions by their actual historical T1 answer gives another view (Table~\ref{tab:history-correctness-summary}; breakdown in Appendix~\ref{app:history-correctness}). In the pooled mathematics results, Native trails Vanilla by 0.53 points after correct T1 answers and 19.72 points after wrong ones. Queries from the current problem determine the attention weights over earlier states. Learning this query-to-history map offers a way to adjust their readout while keeping the stored states fixed.

\section{STAIR Re-addresses Frozen Historical Computation}\label{sec:method}

\begin{figure}[H]
\centering
\includegraphics[width=\linewidth]{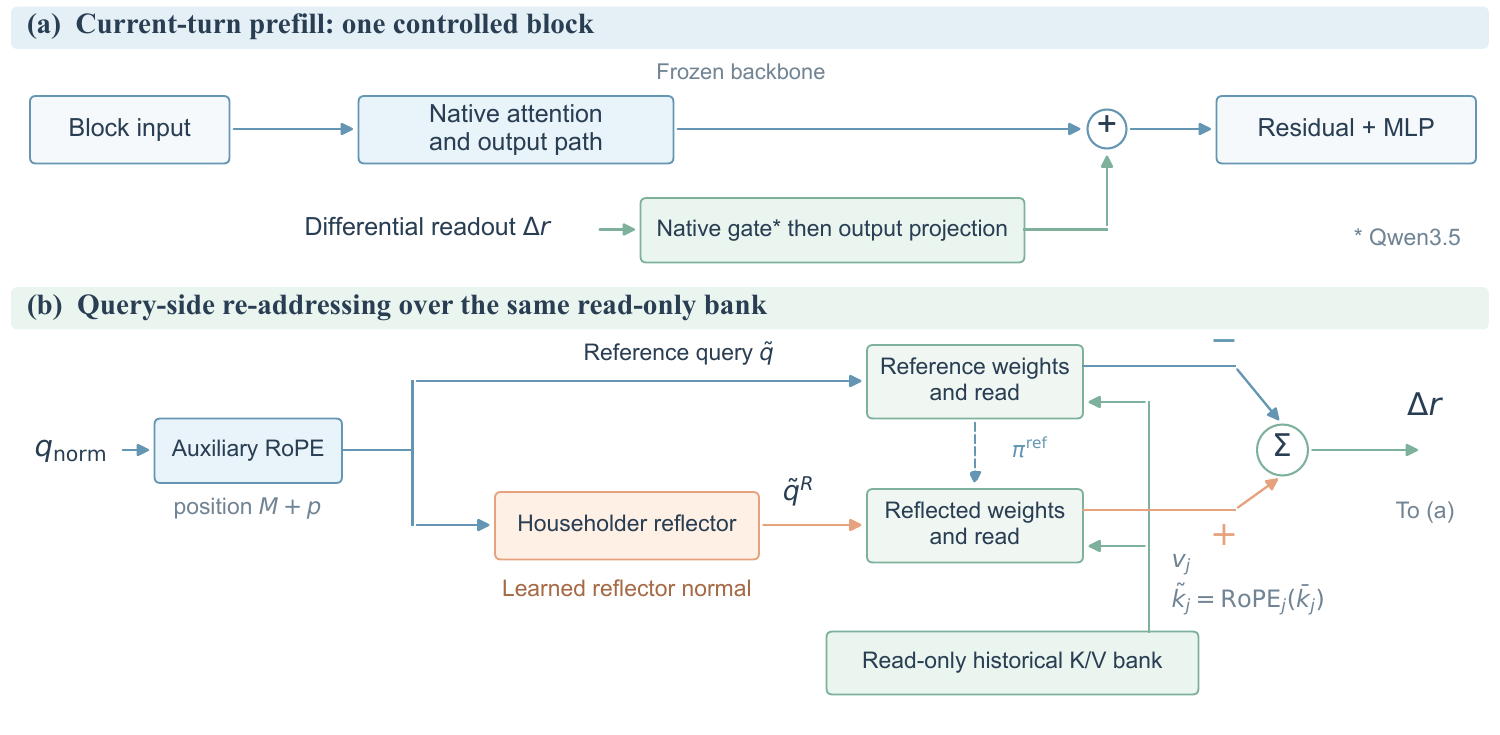}
\caption{STAIR within a controlled block. (a) The change between two reads of the historical bank passes through the frozen output path and is added to the native attention output. Qwen3.5 also uses its native gate. (b) A learned reflection redirects the current query after auxiliary position encoding (RoPE). The original and reflected queries read the same fixed bank; subtracting their readouts gives $\Delta r$. Equation~\ref{eq:method-reflected} specifies the reflected read. Only the reflection directions are learned.}
\label{fig:method}
\end{figure}

To control how the current problem uses earlier computation, STAIR learns how its queries access a separate, fixed historical K/V bank (Figure~\ref{fig:method}). Both Native and STAIR reprocess the visible conversation at each turn. In the thinking models, this text contains earlier questions and post-thinking answers; STAIR's bank additionally retains states from the reasoning that produced them. It makes these states accessible without adding reasoning text to the prompt. Instruct retains the complete assistant answers in its conversation and captures their states in the bank. The auxiliary read acts only on the current prompt suffix, from the user-role header through the assistant generation prefix.

\subsection{A Read-only Historical K/V Bank}\label{sec:method-bank}

Keys determine where a query attends; values supply the content it reads. At turn $t$, controlled layer $\ell$ and K/V head $g$ store the ordered bank
\begin{equation}
\mathcal B_{t,\ell,g}
=\{(\bar k_{\ell,g,j},v_{\ell,g,j})\}_{j=0}^{M_t-1},
\label{eq:method-bank}
\end{equation}
where $M_t$ is the number of stored token positions. Keys are captured after key projection and any native key normalization, before rotary position encoding (RoPE; \citealp{su2024roformer}); paired values are the corresponding value-projection outputs. RoPE rotates queries and keys according to their token positions.

New entries are detached when captured and appended after each turn; existing entries remain unchanged. Appendix~\ref{app:method-capture} specifies the capture boundaries.

The auxiliary branch assigns consecutive positions $0,\ldots,M_t-1$ to bank keys. A current token at position $p$ in the full conversation input, excluding left padding, receives auxiliary position $M_t+p$. The positioned query and keys are
\begin{equation}
\tilde q=\operatorname{RoPE}_{M_t+p}(q_{\mathrm{norm}}),
\qquad
\tilde k_j=\operatorname{RoPE}_{j}(\bar k_j),
\label{eq:method-positions}
\end{equation}
where $q_{\mathrm{norm}}$ is the query after projection and any native query normalization. The native attention path retains its original positions and masks.

\subsection{Query-side Re-addressing}\label{sec:method-addressing}

STAIR changes historical attention weights by reflecting the current query before the auxiliary read. The layerwise diagnostics in Section~\ref{sec:historical-computation} motivated layers $[3,11,19]$ (zero-based), spanning early and intermediate computation. This placement stays fixed across the tested backbones.

Each controlled query head learns a reflector normal $n_{\ell,h}$, a vector perpendicular to the reflection plane. Its Householder reflection \citep{householder1958unitary} preserves query norm by reversing the component along this direction:
\begin{equation}
u_{\ell,h}=\frac{n_{\ell,h}}{\max(\|n_{\ell,h}\|_2,10^{-8})},
\qquad
\tilde q^R=\tilde q-2u_{\ell,h}(u_{\ell,h}^{\top}\tilde q).
\label{eq:method-reflection}
\end{equation}
The learned normal is fixed during inference, while the displacement depends on each query through $u_{\ell,h}^{\top}\tilde q$. The orthogonal query component is unchanged. Reflection follows auxiliary RoPE.

The original positioned query defines a smoothed reference distribution over the bank:
\begin{equation}
\pi_j^{\mathrm{ref}}
=(1-\varepsilon)\operatorname{softmax}_j
\left(\frac{\tilde q^{\top}\tilde k_j}{\sqrt{d_h}}\right)
+\frac{\varepsilon}{M_t},
\qquad \varepsilon=10^{-6},
\label{eq:method-reference}
\end{equation}
where $d_h$ is the head dimension. The reflected query changes the query--key scores. With $c=\tilde q^R-\tilde q$, its distribution reweights the smoothed reference:
\begin{equation}
\pi_j^R
=\frac{\pi_j^{\mathrm{ref}}\exp(c^{\top}\tilde k_j/\sqrt{d_h})}
{\sum_s\pi_s^{\mathrm{ref}}\exp(c^{\top}\tilde k_s/\sqrt{d_h})}.
\label{eq:method-reflected}
\end{equation}
Both distributions normalize over the same bank. The controller determines where to read; the bank supplies the content at those addresses.

\subsection{Differential Readout and Learning}\label{sec:method-learning}

The reference read $r^{\mathrm{ref}}=\sum_j\pi_j^{\mathrm{ref}}v_j$ and reflected read $r^R=\sum_j\pi_j^Rv_j$ differ only in their weights. STAIR writes their difference into the current computation:
\begin{equation}
\Delta r=r^R-r^{\mathrm{ref}}
=\sum_j(\pi_j^R-\pi_j^{\mathrm{ref}})v_j.
\label{eq:method-difference}
\end{equation}
The difference weights sum to zero, so the differential readout captures the change in content induced by redistributing attention over the same bank.

Concatenated head-wise readouts pass through Qwen3.5's native sigmoid gate, when present, and the frozen attention output projection weight. The update is added to the native attention output before the block's residual addition and MLP. The auxiliary branch is disabled after prefill. Decoding uses the modified prefix states along the original path, while new states are captured for later turns.

We train the three main Qwen controllers on their own backbones' responses to DAPO-Math-17k problems \citep{yu2025dapo,dapo2025math17k}, retaining correct and incorrect answers. Responses form four-turn sessions: Native replay of T1 supplies history, and teacher-forced T2--T4 responses supply supervision and new bank entries. As at inference, the controller acts on the current prompt suffix. For target token $y$ and text prefix $s$, the objective combines answer prediction with a Native anchor:

\begin{equation}
\ell(y,s)=-\log p_{\mathrm{STAIR}}(y\mid s)
+\lambda D_{\mathrm{KL}}\!\left(
p_{\mathrm{Native}}(\cdot\mid s)\,\Vert\,
p_{\mathrm{STAIR}}(\cdot\mid s)\right),
\qquad \lambda=1.
\label{eq:method-objective}
\end{equation}
Answer-token losses backpropagate through the modified prompt states to the reflector normals. Their gradients adjust historical attention weights to improve target-token prediction, while the KL term anchors predictions to Native. The frozen backbone transmits these gradients; stored bank entries remain detached across turns. The Native teacher processes the same text independently. Losses are averaged within each response and weighted to balance problems. Only reflector normals are updated (Appendix~\ref{app:method-gradients}--\ref{app:method-objective}).

\section{Experiments}\label{sec:experiments}

\subsection{Experimental Setup}\label{sec:experimental-setup}

Our main study uses Qwen3-4B Instruct \citep{yang2025qwen3,qwen2025instruct2507}, Qwen3.5-4B, and Qwen3.5-9B (both in thinking mode) \citep{qwen2026qwen35,qwen2026qwen35nine}. Each controller trains for one epoch on its backbone's DAPO-Math-17k responses, using 14,806 training and 128 validation problems. We evaluate the final checkpoint on MATH-500 \citep{hendrycks2021math,lightman2023verify}, AIME 2025 \citep{zhang2025aime}, AMC23$^{\dagger}$ \citep{mathaiamc23}, and out-of-domain scientific reasoning on GPQA-Diamond \citep{rein2024gpqa}. Four samples per problem yield Avg@4 (mean correctness) and Pass@4 (at least one correct answer). Paired pointwise 95\% intervals for these evaluations account for question and session clustering (Appendix~\ref{app:paired-intervals}). Appendix~\ref{app:evaluation} specifies training, sampling, and scoring.

Native and STAIR share problem schedules and sample seeds, then continue from their own generated histories; Vanilla answers each problem independently. Shared-history studies at T2 and T4 compare methods at the first historical read and after three completed turns, respectively.

\subsection{Continuous Reasoning}\label{sec:continuous-results}

Table~\ref{tab:continuous-summary} compares mean T2--T4 performance with Native and the no-history Vanilla reference. Figure~\ref{fig:overview}d shows paired differences for the two 4B models; Appendix~\ref{app:continuous-results} gives all turns and intervals.

\begin{table}[H]
\centering
\setlength{\tabcolsep}{3pt}
\renewcommand{\arraystretch}{1.12}
\caption{Continuous reasoning results (\%). Vanilla reports T1; Native and STAIR average T2--T4. Bold marks the better Native/STAIR score, including ties; underlining marks STAIR scores above Vanilla. Paired intervals appear in Appendix~\ref{app:paired-intervals}.}
\label{tab:continuous-summary}
\begin{tabular}{lcccccccc}
\toprule
& \multicolumn{6}{c}{Mathematics} & \multicolumn{2}{c}{Science (OOD)}\\
\cmidrule(lr){2-7}\cmidrule(lr){8-9}
& \multicolumn{2}{c}{MATH-500} & \multicolumn{2}{c}{AIME 2025} & \multicolumn{2}{c}{AMC23$^{\dagger}$} & \multicolumn{2}{c}{GPQA-Diamond}\\
\cmidrule(lr){2-3}\cmidrule(lr){4-5}\cmidrule(lr){6-7}\cmidrule(lr){8-9}
Condition & Avg@4 & Pass@4 & Avg@4 & Pass@4 & Avg@4 & Pass@4 & Avg@4 & Pass@4\\
\midrule
\multicolumn{9}{l}{\textbf{\textit{Qwen3-4B Instruct}}}\\
\rowcolor{vanillarow}Vanilla & 95.00 & 97.20 & 47.50 & 66.67 & 92.65 & 97.06 & 58.59 & 78.79\\
Native & 94.10 & \textbf{97.60} & 34.44 & 47.78 & \textbf{93.38} & 96.08 & 58.00 & 76.94\\
\rowcolor{stairrow}\textbf{STAIR} & \textbf{94.63} & \underline{97.53} & \textbf{38.06} & \textbf{52.22} & \underline{\textbf{93.38}} & \textbf{97.06} & \underline{\textbf{58.75}} & \textbf{77.78}\\
\midrule
\multicolumn{9}{l}{\textbf{\textit{Qwen3.5-4B}}}\\
\rowcolor{vanillarow}Vanilla & 96.00 & 98.60 & 60.83 & 80.00 & 76.47 & 85.29 & 63.64 & 75.76\\
Native & 94.78 & 98.80 & 40.56 & 64.44 & 68.14 & 87.25 & 75.25 & \textbf{89.39}\\
\rowcolor{stairrow}\textbf{STAIR} & \underline{\textbf{96.27}} & \underline{\textbf{99.27}} & \textbf{52.22} & \textbf{77.78} & \textbf{74.51} & \underline{\textbf{90.20}} & \underline{\textbf{75.38}} & \underline{86.87}\\
\midrule
\multicolumn{9}{l}{\textbf{\textit{Qwen3.5-9B}}}\\
\rowcolor{vanillarow}Vanilla & 96.65 & 98.20 & 61.67 & 73.33 & 77.94 & 85.29 & 68.81 & 80.30\\
Native & 96.38 & \textbf{99.20} & 43.33 & 68.89 & 72.55 & 91.18 & \textbf{79.88} & \textbf{89.73}\\
\rowcolor{stairrow}\textbf{STAIR} & \underline{\textbf{96.78}} & \underline{99.13} & \textbf{49.44} & \textbf{73.33} & \textbf{75.74} & \underline{\textbf{92.16}} & \underline{75.93} & \underline{88.55}\\
\bottomrule
\end{tabular}
\par\smallskip\noindent$\dagger$ AMC23 excludes problems overlapping the training set, leaving 34 problems.
\end{table}

The largest gains over Native occur on AIME 2025: 11.67 Avg@4 points for Qwen3.5-4B (paired 95\% interval $[3.16,20.77]$), 6.11 for Qwen3.5-9B, and 3.61 for Instruct. All three improve each later turn's Avg@4 and mean Pass@4 on this benchmark. Both Qwen3.5 models exceed Vanilla in both mean metrics on MATH-500. At the final turn, Qwen3.5-4B exceeds Vanilla in Avg@4 on all four tasks, reaching 64.17\% versus 60.83\% on AIME 2025 (Table~\ref{tab:turns-qwen35_4b_thinking}). On GPQA-Diamond, the two 4B models gain 0.76 and 0.13 mean Avg@4 points over Native. Qwen3.5-9B trails Native by 3.96 points on average; its gap narrows from 7.70 points at T2 to 1.14 at T4 (Table~\ref{tab:turns-qwen35_9b_thinking}).

\begin{samepage}
To examine the broader applicability of learned historical readout, we extend the evaluation across model families and training stages and test transfer to code generation. On AIME 2025, a controller trained for DeepSeek-R1-Distill-Llama-8B \citep{deepseek2025r1llama8b} raises mean T2--T4 Avg@4 from 23.98\% to 28.80\% and Pass@4 from 41.85\% to 43.70\% over Native (Table~\ref{tab:llama-evaluation-seeds}). At 1.5B, we compare a reasoning-distilled model trained by supervised fine-tuning (SFT) with its reinforcement-learning (RL) continuation. With the same SFT-generated T1 histories, their stage-specific controllers raise T2 Avg@4 over Native by 5.28 and 6.39 points, respectively (Appendix~\ref{app:stage-results}). Both AIME studies use three decoding-seed groups, with positive Avg@4 gains in every group for each checkpoint.

\end{samepage}

For code generation, we apply Qwen3.5-9B's math-trained controller unchanged to HumanEval+ \citep{liu2023evalplus}. It raises mean T2--T4 Avg@4 from 86.18\% to 87.91\% and Pass@4 from 93.90\% to 94.31\% over Native, with positive Avg@4 gains at every later turn (Table~\ref{tab:turns-humanevalplus}).

\subsection{Historical Readout from a Shared T1 History}\label{sec:fixed-bank-results}

To isolate the first historical read, we evaluate Qwen3.5-4B on AIME 2025 at T2, sharing Native T1 responses generated separately from the natural rollouts. Exact-token replay builds their banks; all conditions match visible history, current problem, generation seed, and decoding budget.

Table~\ref{tab:shared-t1} compares STAIR with test-time variants and separately trained controls. Random-reflector replaces each head's learned normal with a fixed random direction and retains the same bank read. Direct reflected-read adds $r^R$ in place of $r^R-r^{\mathrm{ref}}$, retaining the learned normals. Bank-free learns 12,288 scales on the current attention output with matched data and updates. K--V pairing permutation shuffles historical values within each K/V head, keeping keys and positions fixed. We apply this permutation at test time alone or throughout controller training and evaluation.

\begin{table}[H]
\centering
\setlength{\tabcolsep}{7pt}
\renewcommand{\arraystretch}{1.12}
\caption{Shared-T1 AIME 2025 results for Qwen3.5-4B at T2 (\%). The 120 shared histories are generated separately from Table~\ref{tab:continuous-summary}'s natural rollouts. Random-reflector and both K--V permutation conditions average three test-time seeds. $\Delta$ is the Avg@4 difference from Native.}
\label{tab:shared-t1}
\begin{tabular}{lccc}
\toprule
Condition & Avg@4 & $\Delta$ (pp) & Pass@4\\
\midrule
Native & 21.67 & 0.00 & 56.67\\
Random-reflector & 37.22 & +15.56 & 63.33\\
Direct reflected-read & 41.67 & +20.00 & 73.33\\
Bank-free & 36.67 & +15.00 & 66.67\\
K--V permutation (test only) & 33.06 & +11.39 & 61.11\\
K--V permutation (train + test) & 31.67 & +10.00 & 58.89\\
\rowcolor{stairrow}\textbf{STAIR} & \textbf{48.33} & \textbf{+26.67} & \textbf{80.00}\\
\bottomrule
\end{tabular}
\end{table}

STAIR gains 26.67 Avg@4 points over Native and 11.67 over Bank-free (95\% interval $[2.50,20.83]$). Random and direct reads also improve on Native, while trailing STAIR. With intact K--V pairs, learned directions outperform random ones by 11.11 points. Permuting values reverses this gap to $-3.89$ points, a 15.00-point interaction (95\% interval $[6.39,24.44]$; Table~\ref{tab:shared-t1-intervals}). Training with permuted pairs yields 31.67\% Avg@4 under test-time permutation, versus 48.33\% for intact STAIR. These comparisons link the learned readout's advantage to historical K--V correspondence.

\subsection{Comparison with History Management}\label{sec:history-management}

At T4, three completed turns provide a longer history for comparing readout with history management. We use Qwen3.5-4B on AIME 2025 with common Native T1--T3 trajectories. Summarization replaces the visible history with a short summary; turn selection retains at most one chosen turn. Sparse access keeps the text intact and restricts historical full-attention reads to the top-ranked 25\% or 50\% of content positions. To compare readout rules on the same bank, we also train a 12,288-parameter reference-read gate that scales the unreflected read, matching STAIR's training data and updates. Appendix~\ref{app:history-management} specifies the controls and the history-management combinations with STAIR.

\begin{table}[!htb]
\centering
\setlength{\tabcolsep}{12pt}
\renewcommand{\arraystretch}{1.12}
\caption{T4 readout and history-management results for Qwen3.5-4B on AIME 2025 (\%). Entries are means over three decoding-seed groups, with standard deviations in subscripts. Each group uses four samples per problem and the same Native T1--T3 histories. Bold marks column maxima.}
\label{tab:history-management}
\begin{tabular}{lrr}
\toprule
Condition & Avg@4 & Pass@4\\
\midrule
Native & \meanstd{46.11}{3.37} & \meanstd{76.67}{5.77}\\
History summarization & \meanstd{67.50}{4.41} & \meanstd{78.89}{1.92}\\
Turn selection & \meanstd{36.39}{1.73} & \meanstd{65.56}{5.09}\\
Sparse-25\% & \meanstd{49.72}{3.47} & \meanstd{78.89}{1.92}\\
Sparse-50\% & \meanstd{51.11}{3.94} & \meanstd{76.67}{6.67}\\
Reference-read gate & \meanstd{57.22}{2.10} & \meanstd{81.11}{5.09}\\
\midrule
\rowcolor{stairrow}\textbf{STAIR} & \meanstd{68.33}{2.50} & \meanstd{\textbf{82.22}}{6.94}\\
\rowcolor{stairrow}STAIR + Summarization & \meanstd{68.61}{3.15} & \meanstd{80.00}{0.00}\\
\rowcolor{stairrow}STAIR + Sparse-25\% & \meanstd{\textbf{70.28}}{2.55} & \meanstd{81.11}{1.92}\\
\rowcolor{stairrow}STAIR + Sparse-50\% & \meanstd{69.72}{2.93} & \meanstd{\textbf{82.22}}{5.09}\\
\bottomrule
\end{tabular}
\end{table}

STAIR leads the standalone methods in both metrics (Table~\ref{tab:history-management}). It gains 22.22 Avg@4 points over Native and 11.11 over the reference-read gate. Its Avg@4 exceeds both controls in every seed group. Combining STAIR with Sparse-25\% or Sparse-50\% reaches 70.28\% and 69.72\% Avg@4, gains of 20.56 and 18.61 points over the respective sparse baselines. Both combinations also raise mean Pass@4. Combined with summarization, STAIR reaches 68.61\% Avg@4 and 80.00\% Pass@4.

\subsection{LoRA Comparison and Joint Training}\label{sec:lora-results}

To compare historical readout and weight adaptation, we evaluate STAIR, LoRA \citep{hu2021lora}, and their joint training on Qwen3-4B Instruct. All runs use the same data, objective, and update budget. On AIME 2025, STAIR gains 3.61 Avg@4 points over Native with 12,288 parameters; LoRA uses 336 times as many (Table~\ref{tab:lora-comparison}). Adding STAIR to LoRA gains a further 3.89 points. Both additions raise Pass@4 by 4.44 points.

\begin{table}[!htb]
\centering
\setlength{\tabcolsep}{6pt}
\renewcommand{\arraystretch}{1.12}
\caption{Qwen3-4B Instruct on AIME 2025 (\%). Scores average T2--T4. $\Delta$ is computed before rounding, relative to Native for STAIR and to LoRA for LoRA + STAIR. Bold marks the highest score and gain for each metric, including ties.}
\label{tab:lora-comparison}
\begin{tabular}{lccccc}
\toprule
& & \multicolumn{2}{c}{Avg@4} & \multicolumn{2}{c}{Pass@4}\\
\cmidrule(lr){3-4}\cmidrule(lr){5-6}
Condition & Trainable params & Score & $\Delta$ (pp) & Score & $\Delta$ (pp)\\
\midrule
Native & 0 & 34.44 & -- & 47.78 & --\\
\rowcolor{stairrow}\textbf{STAIR} & 12,288 & 38.06 & +3.61 & 52.22 & \textbf{+4.44}\\
\midrule
LoRA & 4,128,768 & 39.44 & -- & 58.89 & --\\
\rowcolor{stairrow}\textbf{LoRA + STAIR} & 4,141,056 & \textbf{43.33} & \textbf{+3.89} & \textbf{63.33} & \textbf{+4.44}\\
\bottomrule
\end{tabular}
\end{table}

\subsection{Bank Prefix and Runtime}\label{sec:bank-runtime}

On the shared-T1 inputs, reading the first 2K, 8K, or 32K bank tokens yields 35.00\%, 43.33\%, and 39.17\% Avg@4, versus 48.33\% for the full bank.

In a separate fixed-256-token test on 16 long histories (A800), full-bank access adds a median 0.81 s and 6.60 GiB peak memory. Across four complete AIME 2025 sessions on an A100, STAIR averages 116.00 minutes versus 98.25 for Native, with 17.2\% more output tokens. Appendix~\ref{app:component-runtime} gives both runtime protocols and their measurements.

\newpage
\section{Related Work}\label{sec:related-work}

Memory and caching methods preserve earlier context or reuse its representations. Transformer-XL and Recurrent Memory Transformer carry information across text segments through recurrence \citep{dai2019transformerxl,bulatov2022recurrent}. Memorizing Transformers retrieve past key--value pairs by approximate nearest-neighbor search \citep{wu2022memorizing}. StreamingLLM retains initial attention-sink tokens and a recent-token window \citep{xiao2024streamingllm}. For repeated input text, Prompt Cache precomputes reusable prompt modules \citep{gim2024promptcache}, and CacheBlend selectively recomputes cached text chunks to recover their interactions \citep{yao2025cacheblend}. These cache methods reduce prefill work for text in the current input.

Learned inputs and latent reasoning change how a model computes within a task. Prefix-tuning and soft prompt tuning learn continuous inputs for a frozen model \citep{li2021prefix,lester2021prompt}. Pause tokens add computation before answer generation \citep{goyal2024pause}; Coconut feeds hidden states back as inputs for latent reasoning \citep{hao2025coconut}. A learned coprocessor augments the cache with latent embeddings \citep{liu2025deliberation}, and KV-derived representations guide sampling and fast/slow thinking \citep{xing2026beyond}. Differential Transformer subtracts two attention maps before reading shared values \citep{ye2025differential}. STAIR learns query reflections over states captured from completed problems and writes the resulting read difference during prefill.

In conversations that specify one task incrementally, \citet{laban2026lost} find that early assumptions can persist as further instructions arrive. Our sessions present a complete, independent problem at each turn.

\section{Discussion and Limitations}\label{sec:discussion}

The math-trained Qwen3.5-9B controller improves code generation without further training, while transfer to scientific reasoning varies across models. Reflector directions may favor query--history relationships common in training, motivating readout learning over more varied tasks and histories. The joint LoRA result shows that historical access can be learned alongside backbone adaptation. Pretraining on diverse task switches could fit reflector directions together with backbone representations. Joint supervised fine-tuning or reinforcement learning could then adapt historical readout as reasoning behavior changes.

Our experiments cover four-turn sessions. Longer sessions would increase bank storage and prefill costs. The gains from combining STAIR with sparse access motivate studying selective retention as a way to limit this growth. A retention rule would determine which states remain available, while the learned readout would determine how current queries use them.

\section{Conclusion}\label{sec:conclusion}

After a task switch, history-induced representation changes retain recurring problem-dependent structure. STAIR learns how current queries access separately captured historical K/V states, with the backbone and stored entries fixed. Across four benchmarks on three Qwen models, it gains up to 11.67 percentage points in mean T2--T4 Avg@4 over Native with 12,288 trainable parameters. These findings support treating computation from completed problems as a resource with a learnable readout for subsequent reasoning.

\clearpage
\subsection*{AI use statement}
Generative AI tools assisted with phrasing, grammar, and readability during manuscript preparation. The authors reviewed and revised AI-assisted text and take responsibility for the paper's claims and final manuscript.

\subsection*{Ethics statement}
The study evaluates offline four-turn sessions assembled from the cited public question sets. Earlier assistant responses are generated by the evaluated models; no human participants were recruited, and no private user conversations were collected.

\subsection*{Reproducibility statement}
Appendix~\ref{app:historical-computation} documents the replay design and representation measurements. Appendix~\ref{app:method} specifies state capture, query re-addressing, and training. Appendix~\ref{app:evaluation} gives the data, model, prompt, decoding, and scoring protocols, including the shared-history controls. Appendix~\ref{app:continuous-results} reports per-turn results.

\clearpage
\bibliography{references}
\bibliographystyle{stair_preprint}
\clearpage
\appendix
\raggedbottom
\setlength{\intextsep}{8pt plus 2pt minus 2pt}
\setlength{\textfloatsep}{16pt plus 2pt minus 4pt}
\setlength{\abovecaptionskip}{6pt}
\section{Historical Replay and Response Measurements}\label{app:historical-computation}

\subsection{Replay Construction}\label{app:replay-construction}

The replay histories come from four-turn MATH-500 sessions generated by Qwen3-4B Instruct. Each session contains four distinct problems and four independently sampled paths. The first one, two, or three completed turns of a path form a source history. Each current problem is paired with histories whose source problems exclude it.

Two balanced designs vary coverage along the problem and history axes (Table~\ref{tab:replay-design}). Within each design, the two data groups have disjoint current-problem sets and disjoint history-source pools. The history-coverage design crosses 64 current problems with 128 histories per group. The problem-coverage design expands the current-problem set to 128 and retains 32 histories. Figure~\ref{fig:historical-computation} uses the three-turn problem-coverage condition, comprising 8,192 problem--history cells per model across the two groups. Qwen3.5-4B repeats this condition with its own histories (Appendix~\ref{app:qwen35-replication}).

\begin{table}[htbp]
\centering
\caption{Balanced replay designs for Qwen3-4B Instruct. Histories are sampled paths; sessions count distinct problem orders. Source problems count distinct problems in the first three turns.}
\label{tab:replay-design}
\begin{tabular}{llrrrr}
\toprule
Design & Group & Current problems & Histories & Sessions & Source problems\\
\midrule
History coverage & 1 & 64 & 128 & 33 & 81\\
History coverage & 2 & 64 & 128 & 32 & 83\\
Problem coverage & 1 & 128 & 32 & 25 & 67\\
Problem coverage & 2 & 128 & 32 & 23 & 63\\
\bottomrule
\end{tabular}
\end{table}

Across all three history depths, the history-coverage design contains 49,152 nonempty replays. The problem-coverage design reuses 12,288 cells and adds 12,288, giving 61,440 distinct nonempty Qwen3-4B Instruct replays in total. An empty-history reference is also computed for each of the 256 current problems.

All replays use a frozen backbone in BF16 with FlashAttention2 \citep{dao2023flashattention2}. Each history prefix is computed once and reused read-only across current problems. Replays use batches of 16 current problems, native causal masks, and position IDs following the realized conversation. The empty-history reference retains the system prompt and current-turn template. Across history depths, visible content, historical keys and values, and current-token positions vary together.

History length counts the complete tokenized prefix, including system and role-boundary positions. Median lengths at depths one, two, and three are 1,089.5, 2,196.5, and 3,016.5 tokens in group 1, and 849, 2,473.5, and 4,103 tokens in group 2. Three-turn lengths range from 1,259 to 19,136 tokens in group 1 and from 825 to 11,770 in group 2. Histories sharing any source problem are joined into a connected source block. Each problem-coverage group contains 17 such blocks, which define the source separation used in history pairing and resampling.

\subsection{Attention and State Displacement}\label{app:attention-displacement}

Let $\mathcal T_x$ contain the current user-body tokens, beginning after the user-role header and ending before the turn-closing delimiter. This span includes the shared answer instruction and the problem statement. For Qwen3-4B Instruct, we record outputs from all 36 transformer blocks, indexed from 0 to 35, with hidden width 2,560. The body-mean representation used in Section~\ref{sec:historical-computation} is
\begin{equation}
h_\ell(x\mid H)=\frac{1}{|\mathcal T_x|}\sum_{t\in\mathcal T_x}h_{\ell,t}(x\mid H).
\end{equation}
The corresponding state displacement is defined in Equation~\ref{eq:state-displacement}. Figure~\ref{fig:historical-computation}B reports the median of $\|d_\ell(x,H)\|_2$ over crossed cells. At layer 19, this median is approximately 17.76 in both groups; at layer 35 it is 62.07 and 61.97.

\begin{figure}[t]
\centering
\includegraphics[width=\linewidth]{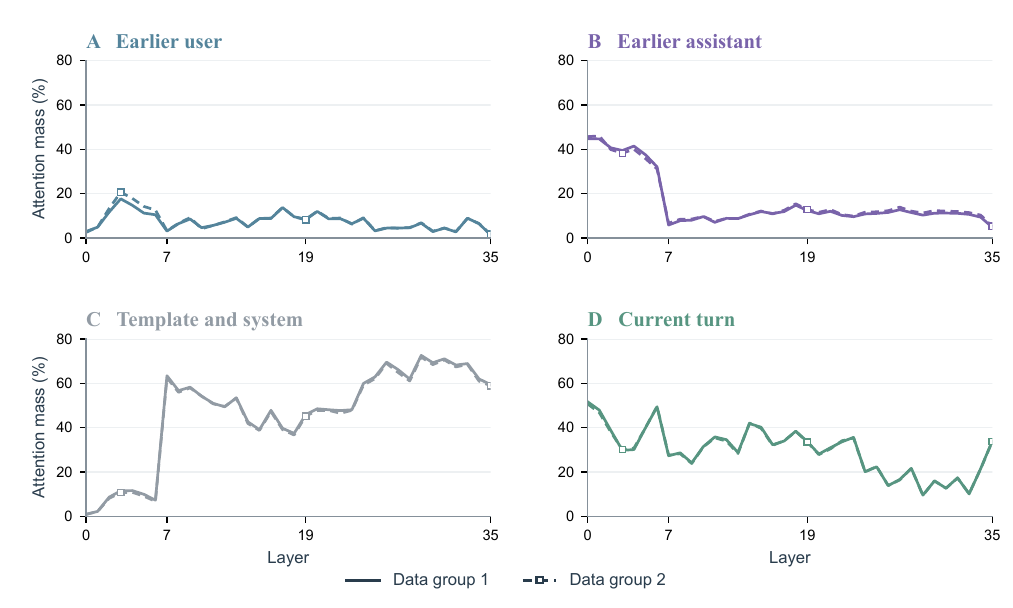}
\caption{Qwen3-4B Instruct Native attention across source categories in the three-turn problem-coverage design. The four panels use common scales. Solid lines denote data group 1; dashed lines with hollow squares denote data group 2. Layer indices are zero-based.}
\label{fig:appendix-attention}
\end{figure}

Let $\mathcal S_H^A$ denote earlier assistant-response body positions and $n_q$ the number of query heads. For native attention probabilities $\alpha_{\ell,h,t,j}$, the attention mass on earlier assistant responses is
\begin{equation}
A_\ell^A(x,H)=\frac{1}{n_q|\mathcal T_x|}
\sum_{h=1}^{n_q}\sum_{t\in\mathcal T_x}\sum_{j\in\mathcal S_H^A}\alpha_{\ell,h,t,j}.
\label{eq:appendix-attention-mass}
\end{equation}
Each head's probabilities are normalized over all causally visible positions. Earlier user bodies, earlier assistant bodies, earlier template and system positions, and current-turn positions form four disjoint source categories. Attention statistics are reconstructed in FP32. For each category, we take the median over current problems within a history, followed by the median over histories. Figure~\ref{fig:appendix-attention} shows their layerwise profiles. Because each category is aggregated separately, the plotted values need not sum to exactly 100\% at a given layer.

\subsection{Interaction Energy and Cross-History Geometry}\label{app:interaction-geometry}

For a fixed layer, history depth, and pooling view, let $n$ and $m$ denote the numbers of current problems and histories. The terms in Equation~\ref{eq:interaction-decomposition} are
\begin{align}
\mu&=\frac{1}{nm}\sum_{x,H}d(x,H), &
a(x)&=\frac{1}{m}\sum_H d(x,H)-\mu,\\
b(H)&=\frac{1}{n}\sum_x d(x,H)-\mu, &
e(x,H)&=d(x,H)-\mu-a(x)-b(H).
\end{align}
Layer subscripts are omitted here. Under the balanced crossing, these components are orthogonal when summed over cells. In particular,
\begin{equation}
\sum_{x,H}\|d(x,H)\|_2^2=nm\|\mu\|_2^2
+m\sum_x\|a(x)\|_2^2+n\sum_H\|b(H)\|_2^2
+\sum_{x,H}\|e(x,H)\|_2^2.
\end{equation}
The interaction energy fraction is the last term divided by total displacement energy. At layer 19 with three-turn histories, it is 0.69\% in group 1 and 0.66\% in group 2.

Neighborhoods are computed in the original 2,560-dimensional interaction-response space using Euclidean distance. Each group's 128 problems are split into four fixed folds, with 32 anchors and 96 candidate problems per fold. Every problem serves as an anchor once. Histories are sorted by realized length, and eight histories are selected at evenly spaced positions. Each is paired with the closest-length history from a different source block.

For an anchor $x$, let $\mathcal N_H^k(x)$ be its $k$ nearest candidate problems under history $H$. The cross-history neighborhood overlap is
\begin{equation}
O_x(H,H';k)=\frac{|\mathcal N_H^k(x)\cap\mathcal N_{H'}^k(x)|}{k}.
\end{equation}
The Question-identity shuffle control permutes anchor identity in the second history within the same fold. Candidate identities, representations, and pairwise distances remain fixed. Overlap is averaged over anchors and folds within each history pair, then over the saved pairs.

\begin{figure}[t]
\centering
\includegraphics[width=\linewidth]{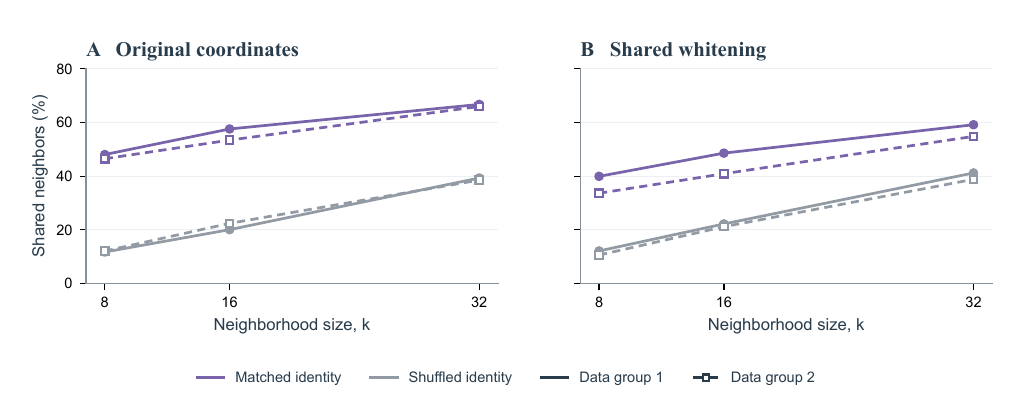}
\caption{Cross-history neighborhood overlap at layer 19 under three-turn histories. Original coordinates use the full-dimensional interaction response; shared whitening uses projected coordinates with a common transform. Purple compares matched problem identities; gray shows the Question-identity shuffle control. Lines connect the three evaluated neighborhood sizes.}
\label{fig:appendix-neighborhood}
\end{figure}

Figure~\ref{fig:appendix-neighborhood} gives the layer-19 results for three neighborhood sizes. A second coordinate view applies a fixed 128-dimensional Gaussian projection followed by shared whitening. Within each fold, candidate coordinates pooled across histories determine a common mean and covariance $\Sigma$. The whitening transform is $(\Sigma+\rho I)^{-1/2}$, with $\rho=10^{-3}\operatorname{tr}(\Sigma)/128$.

At $k=8$, raw state displacement yields overlaps of 73.83\% and 69.82\%. The additive reconstruction yields 100\% in both groups: changing history adds a common translation to all problem representations. The interaction response retains 48.06\% and 46.50\% after those additive components are removed. Matched overlap also exceeds shuffled overlap across the larger neighborhoods and in the whitened coordinates.

\subsection{Attention and State Geometry in Qwen3.5}\label{app:qwen35-replication}

We repeat the three-turn problem-coverage analysis on frozen Qwen3.5-4B, retaining the original question identities, history sources, and two-group assignment. Its own saved Native trajectories yield 8,192 nonempty replays and 256 no-history references from two groups of 128 problems and 32 histories. The saved Native prompts retain earlier questions and the answer text after \texttt{</think>}, providing the ordinary conversation context for this analysis.

\begin{figure}[H]
\centering
\includegraphics[width=\linewidth]{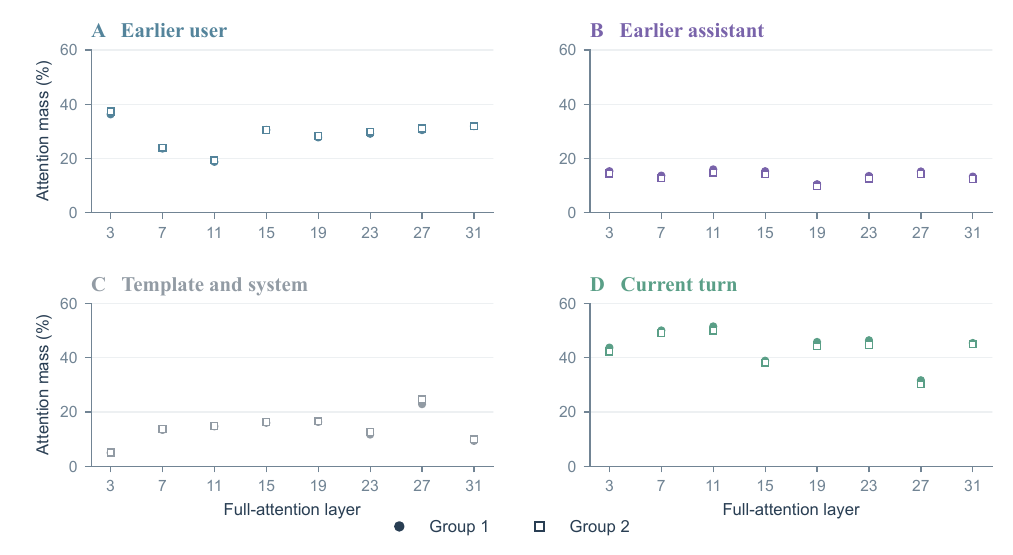}
\caption{Native attention across source categories in Qwen3.5-4B. Points show the eight full-attention layers; filled circles and hollow squares identify groups 1 and 2. Historical assistant bodies contain the visible answer text after thinking. Each category is aggregated separately, so the displayed medians need not sum to 100\%.}
\label{fig:qwen35-attention-sources}
\end{figure}

Native attention mass is measured on the same 8,192 nonempty cells at full-attention layers $[3,7,11,15,19,23,27,31]$. This probability statistic is not defined for the 24 Gated DeltaNet layers. We use the four source categories and the head/token averaging defined in Appendix~\ref{app:attention-displacement}. Probabilities are normalized over all causally visible nonpadding keys. We then take the median over current problems within each history, followed by the median over histories within each group. Figure~\ref{fig:qwen35-attention-sources} shows the resulting source profiles. Earlier assistant bodies receive 9.68--15.94\% of attention mass across the eight layers and two groups.

We average post-block states over the current user-body tokens at all 32 layers, covering both full-attention and Gated DeltaNet blocks. The decomposition, Euclidean neighborhoods, four anchor folds, and source-separated history pairing follow Appendix~\ref{app:interaction-geometry}. We draw 1,000 within-fold anchor-identity permutations, sharing each fold's permutation across its history pairs. These permutations give a pointwise reference range for each layer under the fixed replay design.

\begin{table}[H]
\centering
\setlength{\tabcolsep}{5pt}
\caption{Qwen3.5-4B historical assistant attention mass and state geometry at the three full-attention layers used by STAIR. Layer indices are zero-based. Interaction energy is a percentage of total squared displacement. Matched and shuffled columns report eight-neighbor overlap after removing additive problem and history effects; shuffled values are permutation means.}
\label{tab:qwen35-replication}
\begin{tabular}{ccrrrrr}
\toprule
& & Attention & Median & Interaction & \multicolumn{2}{c}{Neighbor overlap (\%)}\\
\cmidrule(lr){6-7}
Layer & Group & mass (\%) & $\|d_\ell\|_2$ & energy (\%) & Matched & Shuffled\\
\midrule
3 & 1 & 15.28 & 0.587 & 0.99 & 57.70 & 11.51\\
3 & 2 & 14.40 & 0.590 & 1.06 & 53.76 & 11.92\\
\addlinespace
11 & 1 & 15.94 & 1.927 & 1.02 & 55.21 & 12.11\\
11 & 2 & 14.72 & 1.953 & 1.13 & 51.84 & 12.64\\
\addlinespace
19 & 1 & 10.46 & 3.392 & 1.39 & 50.66 & 11.73\\
19 & 2 & 9.68 & 3.419 & 1.57 & 48.22 & 11.79\\
\bottomrule
\end{tabular}
\end{table}

At layers 3, 11, and 19, interaction responses account for 0.99--1.57\% of total displacement energy and retain 48.22--57.70\% neighborhood overlap (Table~\ref{tab:qwen35-replication}). At every layer in both groups, matched overlap exceeds the pointwise 97.5th percentile of the identity-shuffle control. The recurring problem-dependent structure therefore extends to the hybrid backbone.

Replays use the frozen BF16 model with FlashAttention2 and accumulate pooled states in FP32. Each history prefill is reused across current problems, with separate copies of its full-attention K/V cache and DeltaNet convolution and recurrent states for each batch. In eight checked problem--history pairs, mean cached-versus-full-sequence discrepancies were 0.75\%, 0.81\%, and 1.23\% of the measured state displacement at layers 3, 11, and 19, respectively.

The attention observer reconstructs probabilities from native Q/K in FP32 and returns the original FlashAttention2 output unchanged. In four checked inputs, enabling the observer leaves the model outputs identical. Reconstructed $PV$ differs from the native BF16 head output by at most 0.175\% in relative $L_2$ norm, measured before the output gate and projection.

\subsection{Paired Answer Transitions}\label{app:native-accuracy}

The MATH-500 evaluation includes all 500 problems with four sampled responses per problem at each session position, giving 2,000 responses per turn. Each later response is paired with Vanilla T1 by problem and sample identity. The transition analysis uses the scoring rules in Appendix~\ref{app:evaluation-scoring}.

\begin{table}[H]
\centering
\caption{Qwen3-4B Instruct on MATH-500: answer transitions relative to Vanilla T1. C and W denote correct and wrong answers. Net changes are in percentage points.}
\label{tab:native-accuracy}
\begin{tabular}{lrrrrl}
\toprule
Turn & W to C & C to W & C to C & W to W & Net change\\
\midrule
T2 & 37 & 47 & 1,853 & 63 & $-0.50$\\
T3 & 40 & 59 & 1,841 & 60 & $-0.95$\\
T4 & 35 & 60 & 1,840 & 65 & $-1.25$\\
\bottomrule
\end{tabular}
\end{table}

Both transition directions occur at every later turn (Table~\ref{tab:native-accuracy}). Net changes equal the wrong-to-correct count minus the correct-to-wrong count, divided by 2,000 and multiplied by 100.

\clearpage
\subsection{Conditioning on the First History Answer}\label{app:history-correctness}

We group natural continuous sessions by the correctness of the actual T1 response in the same session and sampled trajectory. Native and STAIR share that response, so both conditions use the same grouping. At later turns, each follows its own generated history. The Vanilla reference answers the current problem independently and is matched by problem identity, sample index, and generation seed. Tables~\ref{tab:history-qwen3}--\ref{tab:history-qwen35-9b} report the resulting correct counts and conditional accuracies.

Table~\ref{tab:history-correctness-summary} pools T2--T4 across all three models on MATH-500, AIME 2025, and AMC23$^{\dagger}$. For each T1 group and condition, pooled accuracy is the sum of correct current answers divided by the sum of responses, multiplied by 100. Thus configurations contribute in proportion to their group sizes. The correct-T1 and wrong-T1 pools contain 18,879 and 1,425 later responses, respectively. GPQA-Diamond is reported separately in the tables below. These comparisons describe performance conditional on T1 correctness. Table~\ref{tab:history-correctness-intervals} gives the paired pooled differences and intervals.

\begin{table}[H]
\centering
\setlength{\tabcolsep}{6pt}
\renewcommand{\arraystretch}{1.1}
\caption{Qwen3-4B Instruct: current-answer correct counts and conditional accuracy (\%) by actual T1 correctness. Bold compares Native and STAIR within each group, including ties.}
\label{tab:history-qwen3}
\begin{tabular*}{\linewidth}{@{\extracolsep{\fill}}llcccc@{}}
\toprule
Turn & T1 answer & $n$ & Vanilla & Native & STAIR\\
\midrule
\multicolumn{6}{@{}l}{\textbf{\textit{MATH-500}}}\\
T2 & Correct & 1,900 & 1,808 (95.16) & 1,798 (94.63) & \cellcolor{stairrow}\textbf{1,806 (95.05)} \\
T2 & Wrong & 100 & 92 (92.00) & 92 (92.00) & \cellcolor{stairrow}\textbf{94 (94.00)} \\
T3 & Correct & 1,900 & 1,805 (95.00) & 1,787 (94.05) & \cellcolor{stairrow}\textbf{1,805 (95.00)} \\
T3 & Wrong & 100 & 95 (95.00) & \textbf{94 (94.00)} & \cellcolor{stairrow}\textbf{94 (94.00)} \\
T4 & Correct & 1,900 & 1,804 (94.95) & 1,778 (93.58) & \cellcolor{stairrow}\textbf{1,784 (93.89)} \\
T4 & Wrong & 100 & 96 (96.00) & \textbf{97 (97.00)} & \cellcolor{stairrow}95 (95.00) \\
\addlinespace
\multicolumn{6}{@{}l}{\textbf{\textit{AIME 2025}}}\\
T2 & Correct & 57 & 25 (43.86) & \textbf{20 (35.09)} & \cellcolor{stairrow}19 (33.33) \\
T2 & Wrong & 63 & 32 (50.79) & 24 (38.10) & \cellcolor{stairrow}\textbf{29 (46.03)} \\
T3 & Correct & 57 & 34 (59.65) & 29 (50.88) & \cellcolor{stairrow}\textbf{30 (52.63)} \\
T3 & Wrong & 63 & 23 (36.51) & 10 (15.87) & \cellcolor{stairrow}\textbf{13 (20.63)} \\
T4 & Correct & 57 & 25 (43.86) & 16 (28.07) & \cellcolor{stairrow}\textbf{21 (36.84)} \\
T4 & Wrong & 63 & 32 (50.79) & \textbf{25 (39.68)} & \cellcolor{stairrow}\textbf{25 (39.68)} \\
\addlinespace
\multicolumn{6}{@{}l}{\textbf{\textit{AMC23$^{\dagger}$}}}\\
T2 & Correct & 126 & 116 (92.06) & 118 (93.65) & \cellcolor{stairrow}\textbf{120 (95.24)} \\
T2 & Wrong & 10 & 10 (100.00) & \textbf{10 (100.00)} & \cellcolor{stairrow}9 (90.00) \\
T3 & Correct & 126 & 116 (92.06) & \textbf{117 (92.86)} & \cellcolor{stairrow}114 (90.48) \\
T3 & Wrong & 10 & 10 (100.00) & \textbf{10 (100.00)} & \cellcolor{stairrow}\textbf{10 (100.00)} \\
T4 & Correct & 126 & 116 (92.06) & 116 (92.06) & \cellcolor{stairrow}\textbf{118 (93.65)} \\
T4 & Wrong & 10 & 10 (100.00) & \textbf{10 (100.00)} & \cellcolor{stairrow}\textbf{10 (100.00)} \\
\addlinespace
\multicolumn{6}{@{}l}{\textbf{\textit{GPQA-Diamond}}}\\
T2 & Correct & 464 & 279 (60.13) & \textbf{289 (62.28)} & \cellcolor{stairrow}278 (59.91) \\
T2 & Wrong & 328 & 185 (56.40) & \textbf{193 (58.84)} & \cellcolor{stairrow}185 (56.40) \\
T3 & Correct & 464 & 277 (59.70) & 272 (58.62) & \cellcolor{stairrow}\textbf{276 (59.48)} \\
T3 & Wrong & 328 & 187 (57.01) & 173 (52.74) & \cellcolor{stairrow}\textbf{190 (57.93)} \\
T4 & Correct & 464 & 274 (59.05) & 270 (58.19) & \cellcolor{stairrow}\textbf{272 (58.62)} \\
T4 & Wrong & 328 & 190 (57.93) & 181 (55.18) & \cellcolor{stairrow}\textbf{195 (59.45)} \\
\bottomrule
\end{tabular*}
\par\smallskip\noindent$\dagger$ AMC23 excludes training-overlap problems, leaving 34 problems.
\end{table}

\begin{table}[H]
\centering
\setlength{\tabcolsep}{6pt}
\renewcommand{\arraystretch}{1.1}
\caption{Qwen3.5-4B: current-answer correct counts and conditional accuracy (\%) by actual T1 correctness. Bold compares Native and STAIR within each group, including ties.}
\label{tab:history-qwen35-4b}
\begin{tabular*}{\linewidth}{@{\extracolsep{\fill}}llcccc@{}}
\toprule
Turn & T1 answer & $n$ & Vanilla & Native & STAIR\\
\midrule
\multicolumn{6}{@{}l}{\textbf{\textit{MATH-500}}}\\
T2 & Correct & 1,920 & 1,848 (96.25) & 1,768 (92.08) & \cellcolor{stairrow}\textbf{1,813 (94.43)} \\
T2 & Wrong & 80 & 72 (90.00) & 42 (52.50) & \cellcolor{stairrow}\textbf{59 (73.75)} \\
T3 & Correct & 1,920 & 1,844 (96.04) & 1,864 (97.08) & \cellcolor{stairrow}\textbf{1,875 (97.66)} \\
T3 & Wrong & 80 & 76 (95.00) & 67 (83.75) & \cellcolor{stairrow}\textbf{74 (92.50)} \\
T4 & Correct & 1,920 & 1,844 (96.04) & 1,873 (97.55) & \cellcolor{stairrow}\textbf{1,878 (97.81)} \\
T4 & Wrong & 80 & 76 (95.00) & 73 (91.25) & \cellcolor{stairrow}\textbf{77 (96.25)} \\
\addlinespace
\multicolumn{6}{@{}l}{\textbf{\textit{AIME 2025}}}\\
T2 & Correct & 73 & 47 (64.38) & 29 (39.73) & \cellcolor{stairrow}\textbf{38 (52.05)} \\
T2 & Wrong & 47 & 26 (55.32) & 7 (14.89) & \cellcolor{stairrow}\textbf{13 (27.66)} \\
T3 & Correct & 73 & 47 (64.38) & \textbf{42 (57.53)} & \cellcolor{stairrow}41 (56.16) \\
T3 & Wrong & 47 & 26 (55.32) & 11 (23.40) & \cellcolor{stairrow}\textbf{19 (40.43)} \\
T4 & Correct & 73 & 40 (54.79) & 40 (54.79) & \cellcolor{stairrow}\textbf{49 (67.12)} \\
T4 & Wrong & 47 & 33 (70.21) & 17 (36.17) & \cellcolor{stairrow}\textbf{28 (59.57)} \\
\addlinespace
\multicolumn{6}{@{}l}{\textbf{\textit{AMC23$^{\dagger}$}}}\\
T2 & Correct & 104 & 76 (73.08) & 62 (59.62) & \cellcolor{stairrow}\textbf{67 (64.42)} \\
T2 & Wrong & 32 & 28 (87.50) & 9 (28.12) & \cellcolor{stairrow}\textbf{15 (46.88)} \\
T3 & Correct & 104 & 74 (71.15) & 84 (80.77) & \cellcolor{stairrow}\textbf{90 (86.54)} \\
T3 & Wrong & 32 & 30 (93.75) & 16 (50.00) & \cellcolor{stairrow}\textbf{25 (78.12)} \\
T4 & Correct & 104 & 79 (75.96) & 86 (82.69) & \cellcolor{stairrow}\textbf{89 (85.58)} \\
T4 & Wrong & 32 & 25 (78.12) & \textbf{21 (65.62)} & \cellcolor{stairrow}18 (56.25) \\
\addlinespace
\multicolumn{6}{@{}l}{\textbf{\textit{GPQA-Diamond}}}\\
T2 & Correct & 504 & 316 (62.70) & 387 (76.79) & \cellcolor{stairrow}\textbf{396 (78.57)} \\
T2 & Wrong & 288 & 188 (65.28) & 209 (72.57) & \cellcolor{stairrow}\textbf{218 (75.69)} \\
T3 & Correct & 504 & 303 (60.12) & \textbf{369 (73.21)} & \cellcolor{stairrow}361 (71.63) \\
T3 & Wrong & 288 & 201 (69.79) & 227 (78.82) & \cellcolor{stairrow}\textbf{231 (80.21)} \\
T4 & Correct & 504 & 339 (67.26) & \textbf{395 (78.37)} & \cellcolor{stairrow}382 (75.79) \\
T4 & Wrong & 288 & 165 (57.29) & 201 (69.79) & \cellcolor{stairrow}\textbf{203 (70.49)} \\
\bottomrule
\end{tabular*}
\par\smallskip\noindent$\dagger$ AMC23 excludes training-overlap problems, leaving 34 problems.
\end{table}

\begin{table}[H]
\centering
\setlength{\tabcolsep}{6pt}
\renewcommand{\arraystretch}{1.1}
\caption{Qwen3.5-9B: current-answer correct counts and conditional accuracy (\%) by actual T1 correctness. Bold compares Native and STAIR within each group, including ties.}
\label{tab:history-qwen35-9b}
\begin{tabular*}{\linewidth}{@{\extracolsep{\fill}}llcccc@{}}
\toprule
Turn & T1 answer & $n$ & Vanilla & Native & STAIR\\
\midrule
\multicolumn{6}{@{}l}{\textbf{\textit{MATH-500}}}\\
T2 & Correct & 1,933 & 1,874 (96.95) & 1,850 (95.71) & \cellcolor{stairrow}\textbf{1,857 (96.07)} \\
T2 & Wrong & 67 & 59 (88.06) & 36 (53.73) & \cellcolor{stairrow}\textbf{46 (68.66)} \\
T3 & Correct & 1,933 & 1,867 (96.59) & \textbf{1,892 (97.88)} & \cellcolor{stairrow}1,884 (97.47) \\
T3 & Wrong & 67 & 66 (98.51) & 54 (80.60) & \cellcolor{stairrow}\textbf{58 (86.57)} \\
T4 & Correct & 1,933 & 1,868 (96.64) & 1,892 (97.88) & \cellcolor{stairrow}\textbf{1,900 (98.29)} \\
T4 & Wrong & 67 & 65 (97.01) & 59 (88.06) & \cellcolor{stairrow}\textbf{62 (92.54)} \\
\addlinespace
\multicolumn{6}{@{}l}{\textbf{\textit{AIME 2025}}}\\
T2 & Correct & 74 & 52 (70.27) & 47 (63.51) & \cellcolor{stairrow}\textbf{49 (66.22)} \\
T2 & Wrong & 46 & 22 (47.83) & 10 (21.74) & \cellcolor{stairrow}\textbf{13 (28.26)} \\
T3 & Correct & 74 & 50 (67.57) & \textbf{41 (55.41)} & \cellcolor{stairrow}\textbf{41 (55.41)} \\
T3 & Wrong & 46 & 24 (52.17) & 6 (13.04) & \cellcolor{stairrow}\textbf{16 (34.78)} \\
T4 & Correct & 74 & 41 (55.41) & \textbf{39 (52.70)} & \cellcolor{stairrow}38 (51.35) \\
T4 & Wrong & 46 & 33 (71.74) & 13 (28.26) & \cellcolor{stairrow}\textbf{21 (45.65)} \\
\addlinespace
\multicolumn{6}{@{}l}{\textbf{\textit{AMC23$^{\dagger}$}}}\\
T2 & Correct & 106 & 80 (75.47) & 79 (74.53) & \cellcolor{stairrow}\textbf{82 (77.36)} \\
T2 & Wrong & 30 & 26 (86.67) & 12 (40.00) & \cellcolor{stairrow}\textbf{15 (50.00)} \\
T3 & Correct & 106 & 76 (71.70) & 82 (77.36) & \cellcolor{stairrow}\textbf{87 (82.08)} \\
T3 & Wrong & 30 & 30 (100.00) & 17 (56.67) & \cellcolor{stairrow}\textbf{22 (73.33)} \\
T4 & Correct & 106 & 82 (77.36) & \textbf{88 (83.02)} & \cellcolor{stairrow}86 (81.13) \\
T4 & Wrong & 30 & 24 (80.00) & \textbf{18 (60.00)} & \cellcolor{stairrow}17 (56.67) \\
\addlinespace
\multicolumn{6}{@{}l}{\textbf{\textit{GPQA-Diamond}}}\\
T2 & Correct & 545 & 376 (68.99) & \textbf{442 (81.10)} & \cellcolor{stairrow}409 (75.05) \\
T2 & Wrong & 247 & 169 (68.42) & \textbf{190 (76.92)} & \cellcolor{stairrow}162 (65.59) \\
T3 & Correct & 545 & 359 (65.87) & \textbf{432 (79.27)} & \cellcolor{stairrow}417 (76.51) \\
T3 & Wrong & 247 & 186 (75.30) & \textbf{201 (81.38)} & \cellcolor{stairrow}192 (77.73) \\
T4 & Correct & 545 & 381 (69.91) & 447 (82.02) & \cellcolor{stairrow}\textbf{449 (82.39)} \\
T4 & Wrong & 247 & 164 (66.40) & \textbf{186 (75.30)} & \cellcolor{stairrow}175 (70.85) \\
\bottomrule
\end{tabular*}
\par\smallskip\noindent$\dagger$ AMC23 excludes training-overlap problems, leaving 34 problems.
\end{table}

\clearpage
\subsection{Historical Reads and Downstream MLP Responses}\label{app:read-prediction}

To examine how historical attention relates to later computation, we form the value-weighted read from earlier assistant positions:
\begin{equation}
r_\ell^A(x,H)=\frac{1}{|\mathcal T_x|}\sum_{t\in\mathcal T_x}
W_{O,\ell}\operatorname{Concat}_h\left(\sum_{j\in\mathcal S_H^A}
\alpha_{\ell,h,t,j}v_{\ell,g(h),j}\right),
\end{equation}
where $g(h)$ maps query heads to their grouped key/value head and $W_{O,\ell}$ is the frozen output projection. Earlier-user and template/system reads use the corresponding source positions.

\begin{figure}[H]
\centering
\includegraphics[width=\linewidth]{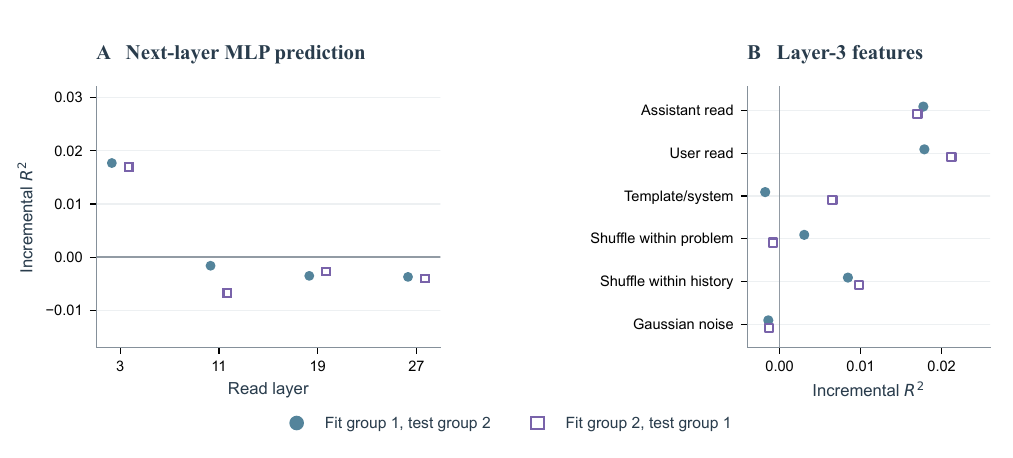}
\caption{Historical reads and next-layer MLP prediction. \textbf{A:} Incremental test $R^2$ from adding the earlier-assistant read at four measured layers. \textbf{B:} Point estimates for equal-width feature additions at layer 3. Shuffling permutes assistant reads across histories within a problem, or across problems within a history, separately in fit and test groups. Colors and markers identify the two data-group exchange directions.}
\label{fig:appendix-read-prediction}
\end{figure}

We test how much this read adds to a predictor of the next layer's MLP response. Let $u_\ell$ be the mean-pooled block input and $z_\ell$ the mean-pooled MLP output. The target is the full-dimensional difference $z_{\ell+1}(x\mid H)-z_{\ell+1}(x\mid\varnothing)$. The base feature contains 64-dimensional projections of $u_\ell(x\mid\varnothing)$ and its history-conditioned difference, log history length, and the three historical source-category attention masses:
\begin{equation}
f_0(x,H)=\left[Pu_\ell(x\mid\varnothing),\;P\Delta u_\ell(x,H),\;\log(1+|H|),\;A_\ell^U,\;A_\ell^A,\;A_\ell^F\right].
\end{equation}
Here $P$ is a fixed Gaussian projection, and $F$ denotes template/system positions. Adding $Pr_\ell^A$ expands the feature from 132 to 196 dimensions. Pooled tensors are stored in FP32; regression uses FP64.

Multi-output ridge regression \citep{hoerl1970ridge} is fitted on one data group and evaluated on the other, with both exchange directions reported. Feature standardization uses only the fit group, and the ridge coefficient equals the number of fit cells. The incremental score is $\Delta R^2=(\mathrm{SSE}_0-\mathrm{SSE}_{\mathrm{read}})/\mathrm{TSS}$ on the held-out group.

The positive increment is concentrated at layer 3 (Figure~\ref{fig:appendix-read-prediction}A): $\Delta R^2$ is 0.0177 when fitting group 1 and testing group 2, and 0.0170 in the reverse direction. At that layer, equal-width earlier-user reads carry comparable predictive information (Figure~\ref{fig:appendix-read-prediction}B). Shuffling assistant reads across histories within each problem reduces the increment to 0.00304 and $-0.00087$. These measurements describe an early association between historical reads and subsequent MLP responses.

\clearpage
\section{STAIR Implementation and Training}\label{app:method}

\subsection{State Capture and Bank Accumulation}\label{app:method-capture}

At each controlled attention layer, the historical K/V bank stores keys after projection and any native key normalization and values after value projection. Keys are captured before RoPE. The bank retains K/V heads in their original order. Under grouped-query attention, several query heads share one K/V head \citep{ainslie2023gqa}; each uses its corresponding stored head. Captured tensors are detached immediately and concatenated along the token dimension in turn order.

For the thinking backbones, the captured span begins at the start of the generated sequence and ends at the start of the last \texttt{</think>} marker. Ordinary history carries the final-answer text after that marker. If the marker is absent, the ordinary assistant message is empty and capture covers the generated prefix that has actually undergone a forward pass. Instruct models retain the assistant-answer body in ordinary history and capture its processed states for the bank. A token sampled at the final decoding step has no captured key or value until it is subsequently processed by the model; this boundary also applies to responses ending at the length limit.

Training constructs banks during teacher-forced replay of the reorganized sessions. T1 uses the Native path; subsequent turns capture states from the student path. Capture is disabled during Native teacher evaluation. For each turn with a successor, the new entries are appended to the existing bank. Earlier entries keep the values captured when they were produced. Evaluation captures states during decoding. For reused T1 responses, we feed the saved tokens through the frozen backbone to construct the initial bank.

The ordinary decoding cache is local to a turn. At evaluation, each backbone re-prefills the full visible conversation at the start of each turn. The resulting ordinary cache serves that turn's autoregressive decoding. The auxiliary historical bank persists separately across turns, retaining states captured during their original generation even as the ordinary context is processed again.

\subsection{Auxiliary Coordinates and Exact Reads}\label{app:method-reads}

The concatenated bank receives auxiliary positions $0,\ldots,M_t-1$. A current token uses $M_t+p$, with $p$ measured in the full ordinary conversation input after removing left padding. The auxiliary branch applies the model's RoPE configuration to these positions, then reflects the positioned query as in Equation~\ref{eq:method-reflection}. Reflector normals are normalized in FP32 using a denominator floor of $10^{-8}$.

Training and evaluation use temperature one and reference smoothing $\varepsilon=10^{-6}$. With $c=\tilde q^R-\tilde q$, we center the logit change using the reference-weighted key mean:
\begin{equation}
\bar k_{\mathrm{ref}}=\sum_j\pi_j^{\mathrm{ref}}\tilde k_j,
\qquad
\pi_j^R=\operatorname{softmax}_j\left(
\log\pi_j^{\mathrm{ref}}+
\frac{c^\top(\tilde k_j-\bar k_{\mathrm{ref}})}{\sqrt{d_h}}
\right).
\label{eq:method-centered-read}
\end{equation}
For a fixed query, the centering term is constant across bank positions and cancels under softmax. This yields the reweighting in Equation~\ref{eq:method-reflected}, including its smoothed reference distribution.

\subsection{Output Path and Prefill Scope}\label{app:method-prefill}

Let $z$ concatenate the head-wise differential readouts at a controlled position. The auxiliary contribution to the attention output is
\begin{equation}
\Delta o = W_O z
\quad\text{(Qwen3)},\qquad
\Delta o = W_O\bigl(g\odot z\bigr)
\quad\text{(Qwen3.5)},
\label{eq:method-output}
\end{equation}
where $g$ is the native sigmoid output gate produced by the query projection. Its parameters and $W_O$ remain frozen. The auxiliary projection uses the output weight alone; the native path retains the original output bias. The runtime adds $\Delta o$ to the native self-attention output before the block performs residual addition and its MLP.

\begin{figure}[htbp]
\centering
\includegraphics[width=\linewidth]{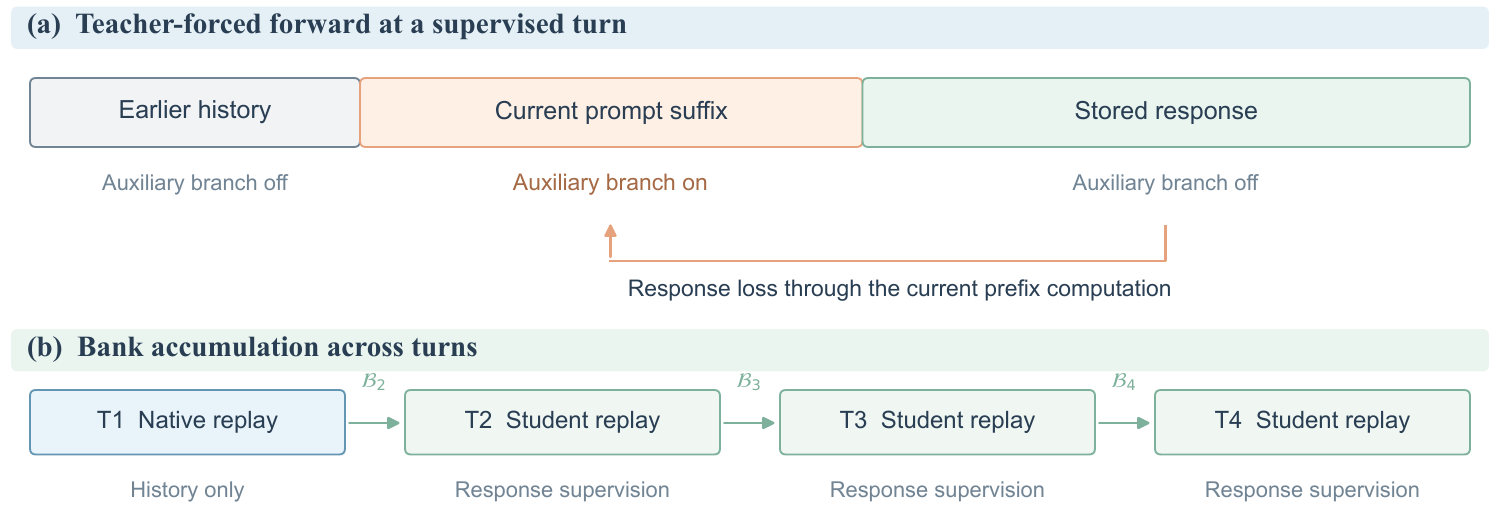}
\caption{Training scope and bank accumulation. (a) The auxiliary branch acts on the current prompt suffix; response losses propagate through the current prefix computation. (b) T1 provides history, and T2--T4 receive response supervision. Each $\mathcal B_t$ contains detached states from all preceding turns. New entries come from the turn's replay and are appended for its successor.}
\label{fig:method-training}
\end{figure}

For Qwen, the control span starts at the last \texttt{<|im\_start|>user} header and ends at the end of the prompt. It includes the user body, intervening template tokens, and assistant generation prefix. Qwen3.5 prompts also supply the opening \texttt{<think>} marker within this span. Earlier history positions and target-response positions have the auxiliary branch disabled (Figure~\ref{fig:method-training}(a)). At inference, the runtime disables the branch after prompt prefill and continues decoding from the resulting states.

\subsection{Session Replay and Gradient Flow}\label{app:method-gradients}

Each main Qwen model's response pool is generated with that same model through SGLang \citep{zheng2024sglang} on problems from the training split of \texttt{BytedTsinghua-SIA/DAPO-Math-17k} \citep{dapo2025math17k}. Rollout settings appear in Appendix~\ref{app:training-settings}. Each uses 14,806 training problems and 128 validation problems. Responses are retained irrespective of correctness or completion status; correctness annotations carry no weight in the training objective. Some problems appear in multiple source records. Their responses remain grouped under the same problem identity. Appendix~\ref{app:llama-extension} describes the Llama response pool, and Appendix~\ref{app:stage-protocol} specifies the shared supervision for the training-stage comparison.

Problems are grouped and matched to exclusion lists by normalized question text, removing fixed prompt prefixes and applying Unicode NFKC normalization, lowercasing, and removal of whitespace and selected LaTeX formatting. All three Qwen training pools use exclusion lists covering MATH-500, AIME 2025, and GPQA-Diamond. The 9B pool additionally excludes the 34-problem AMC23 evaluation subset. Six AMC23 problems retained in all three Qwen training pools are excluded from evaluation, as specified in Appendix~\ref{app:evaluation-scoring}.

Validation is split at the problem level. For the Qwen response pools, the four recorded responses within each source record are randomly assigned one to each turn column. Columns are then shuffled independently, with swaps resolving repeated problem identities within a session. Each recorded response appears once, and each session contains four distinct problems.

Each training session contains four recorded problem--response pairs. T1 supplies the initial visible history and bank. For T2--T4, the student processes the prompt and recorded response in a causal teacher-forced forward pass. The control mask selects only the current-turn prompt suffix. Prediction targets cover the complete stored response, including reasoning, final answer, and any stored end markers. Incomplete responses contribute their existing tokens. The output at the final prompt position predicts the first response token; subsequent response positions predict the remaining targets.

The current-turn prefix computation retains its gradient graph. Response losses therefore reach the reflector normals through their dependence on the controlled prefix states. Captured bank tensors are detached, so later turns treat the accumulated bank as fixed input (Figure~\ref{fig:method-training}(b)).

\begin{samepage}
The Native teacher independently processes the same ordinary history, prompt, and teacher-forced response prefix under disabled gradient recording. Where supported, earlier-history Native KV states can be reused as a detached prefix, ending before the current user message. Cross-turn Native KV reuse is disabled for the Qwen3.5 hybrid architecture, and its student forward runs without a KV cache. These choices preserve the current-turn gradient path while keeping the historical bank detached.

\end{samepage}

\begin{samepage}
\subsection{Supervision and Parameter Updates}\label{app:method-objective}

The token objective in Equation~\ref{eq:method-objective} uses full-vocabulary $D_{\mathrm{KL}}(p_{\mathrm{Native}}\Vert p_{\mathrm{STAIR}})$ with coefficient one. The Native distribution is detached, while the student loss preserves gradients to the student hidden states.

For a supervised response $r$ of length $T_r$, let $w_r$ be its precomputed problem-balancing weight. An optimizer update over a set $\mathcal U$ of supervised responses uses
\begin{equation}
\mathcal L_{\mathcal U}
=\frac{1}{|\mathcal U|}\sum_{r\in\mathcal U}
w_r\left[\frac{1}{T_r}\sum_{a=1}^{T_r}\ell(y_{r,a},s_{r,a})\right].
\label{eq:method-weighted-objective}
\end{equation}
\end{samepage}
For the Qwen session construction, with $Q$ training problems and $S$ supervised responses in the full collection, a problem with $C_q$ recorded candidates contributes $3C_q/4$ supervised responses. Each receives weight $w_r=(S/Q)/(3C_q/4)$. This gives every problem equal total weight while retaining all source rows. Update-level normalization uses the actual number of supervised responses across workers. T1 contributes history without a supervised response loss.

Only reflector normals receive optimizer updates. All tested backbones control every query head at zero-based layer indices $[3,11,19]$. Each layer--head pair has its own normal. The model-specific parameter count is \mbox{$P=\sum_{\ell\in\mathcal L}H_{q,\ell}d_{h,\ell}$} (Table~\ref{tab:controller-configuration}). Backbone parameters, the vocabulary head, any native output gate, and the attention output projection remain frozen.

In Qwen3.5, all three selected positions are native full-attention layers \citep{qwen2026qwen35,qwen2026qwen35nine}. STAIR reuses the keys and values produced at these layers. Gated DeltaNet layers \citep{yang2025gateddelta} retain their original computation without an auxiliary STAIR branch.

\begin{table}[htbp]
\centering
\caption{Controller configuration. All query heads are controlled at layers $[3,11,19]$ (zero-based). Normals are separate for each layer and head.}
\label{tab:controller-configuration}
\begin{tabular}{lrrrr}
\toprule
Model & Query heads/layer & $d_h$ & Init. std. & Parameters\\
\midrule
Qwen3-4B Instruct & 32 & 128 & $1/\sqrt{128}$ & 12,288\\
Qwen3.5-4B & 16 & 256 & $1/16$ & 12,288\\
Qwen3.5-9B & 16 & 256 & $1/16$ & 12,288\\
R1-Distill-Llama-8B & 32 & 128 & $1/\sqrt{128}$ & 12,288\\
R1-Distill-Qwen-1.5B & 12 & 128 & $1/\sqrt{128}$ & 4,608\\
DeepScaleR-1.5B & 12 & 128 & $1/\sqrt{128}$ & 4,608\\
\bottomrule
\end{tabular}
\end{table}

At the start of training, normal coordinates are sampled independently as $n_{\ell,h,j}\sim\mathcal N(0,1/d_h)$. The forward pass normalizes each vector as in Equation~\ref{eq:method-reflection}.

\clearpage
\section{Training and Evaluation Protocols}\label{app:evaluation}

\subsection{Training Settings}\label{app:training-settings}

The Qwen backbones are \texttt{Qwen/Qwen3-4B-Instruct-2507}, \texttt{Qwen/Qwen3.5-4B}, and \texttt{Qwen/Qwen3.5-9B}. Their training rollouts contain four responses per problem, with top-$k$ 20, min-$p$ 0, and repetition penalty 1. Instruct uses temperature 0.7, top-$p$ 0.8, presence penalty 0, and a 16,384-token output limit. Both Qwen3.5 models use temperature 1.0, top-$p$ 0.95, and presence penalty 1.5, with generation limits of 32,768 tokens for 4B and 131,072 for 9B.

Training problems follow the instruction ``Please reason step by step, and put your final answer within \texttt{\textbackslash boxed\{\}}.'', followed by a blank line and the problem text. The model-specific rollouts have no system message. Instruct uses the non-thinking chat template; both Qwen3.5 models use the thinking template with an assistant prefix of \texttt{<think>} followed by a newline. Training replay uses the same thinking template and prefix.

STAIR uses AdamW \citep{loshchilov2019adamw} with learning rate 0.003, zero weight decay, and no scheduler or warmup. Training runs for one epoch in BF16 with FlashAttention2 \citep{dao2023flashattention2}. A full update contains eight four-turn sessions and 24 supervised responses at T2--T4; T1 supplies history. The objective combines CE with full-vocabulary $D_{\mathrm{KL}}(p_{\mathrm{Native}}\Vert p_{\mathrm{STAIR}})$ at coefficient and temperature one. Response-level token averaging and problem-balancing weights follow Appendix~\ref{app:method-objective}.

The scheduled update counts are 1,851 for Qwen3-4B Instruct, 1,941 for Qwen3.5-4B, and 1,940 for Qwen3.5-9B. Checkpoints are selected at the final scheduled update, independently of validation or test performance. Each of these models trains 12,288 reflector parameters at zero-based layer indices $[3,11,19]$.

The LoRA baseline uses Qwen3-4B Instruct with rank 2, alpha 4, and dropout 0. Adapters cover \texttt{q\_proj}, \texttt{k\_proj}, \texttt{v\_proj}, \texttt{o\_proj}, \texttt{gate\_proj}, \texttt{up\_proj}, and \texttt{down\_proj} in all 36 layers, leaving embeddings, the vocabulary head, and biases unchanged. This configuration has 4,128,768 trainable parameters, 336 times the STAIR parameter count. It uses AdamW with learning rate $2\times10^{-4}$ and zero weight decay, with 56 linear warmup steps followed by cosine decay to zero over the 1,851-step schedule. Training data, session construction, supervision, objective, and update budget match Instruct STAIR. Parameter count and optimization schedule differ; compute budgets have not been measured.

LoRA + STAIR initializes both modules afresh and trains them jointly, for a total of 4,141,056 parameters. Each module retains its configuration, learning rate, and schedule above. Data, session construction, objective, and the 1,851-step budget match the standalone Instruct runs. The Native teacher has both modules disabled. At evaluation, LoRA and LoRA + STAIR each generate their own T1 histories. In the joint model, LoRA acts at T1; STAIR begins at T2 once the historical bank is available.

\subsection{Sampling and Matched Conditions}\label{app:evaluation-sampling}

For a benchmark with $N$ problems, we construct $N$ four-turn sessions. Each turn column is a permutation of the benchmark: every problem occurs once at T1, once at T2, once at T3, and once at T4, across different sessions. The four problems within a session are distinct. The same problem can therefore supply history in one session and be evaluated at a later position in another. These assignments are fixed across models and conditions.

Each session is run along four sampled trajectories. Trajectory $s$ at a later turn retains the earlier responses from that same trajectory. Seeds depend on benchmark, problem identity, and sample index, so the same problem--sample pair uses the same seed across positions and conditions. Vanilla evaluates every problem independently. Its saved responses also supply the shared T1 reference for Native and STAIR where reused.

Each problem has four sampled responses per turn. Native and STAIR use matched problems, reference answers, turn assignments, and sample seeds, with each condition continuing from its own generated history. Generation uses Transformers \citep{wolf2020transformers} with FlashAttention2; Qwen3.5-9B runs in BF16. Table~\ref{tab:evaluation-decoding} gives the recorded decoding settings for the three Qwen models. All use $\texttt{min\_p}=0$, repetition penalty 1, and the corresponding official mode template.

\begin{table}[htbp]
\centering
\caption{Qwen decoding settings. Output limits count newly generated tokens per problem.}
\label{tab:evaluation-decoding}
\begin{tabular}{lrrrrl}
\toprule
Model & Temp. & Top-$p$ & Top-$k$ & Presence penalty & Output limit\\
\midrule
Qwen3-4B Instruct & 0.7 & 0.8 & 20 & 0 & 16,384\\
Qwen3.5-4B & 1.0 & 0.95 & 20 & 1.5 & 81,920 / 32,768\\
Qwen3.5-9B & 1.0 & 0.95 & 20 & 1.5 & 81,920 / 32,768\\
\bottomrule
\end{tabular}
\end{table}

For both Qwen3.5 models, the 81,920-token evaluation limit applies to MATH-500 and AIME 2025; GPQA-Diamond and AMC23 use 32,768.

Evaluation uses the official chat templates without an additional system message, with thinking disabled for Instruct and enabled for Qwen3.5. Mathematical prompts read ``Solve the following problem. Show your reasoning, and put the final answer inside \texttt{\textbackslash boxed\{\}}.'', followed on a new line by ``Problem: \texttt{\{problem\}}''. GPQA prompts read ``Answer the following multiple-choice question. Reason carefully. Put your final answer, consisting of only the choice letter, inside \texttt{\textbackslash boxed\{\}}, for example \texttt{\textbackslash boxed\{C\}}.'', followed by ``Question:'' and the problem on separate lines. The GPQA problem includes the question and its fixed A--D choice ordering.

\subsection{Answer Scoring}\label{app:evaluation-scoring}

For mathematical tasks, the scorer extracts the complete answer associated with the last \texttt{\textbackslash boxed} occurrence in each response. Equivalence checks account for the requested answer format, including angle units, percentages, ordinals, and matrices. Integer-answer problems accept equivalent decimal or fractional representations and reject ambiguous multiple answers.

For GPQA-Diamond, each question has a fixed pseudorandom ordering of its four choices, shared across conditions. Scoring extracts the last boxed answer, removes supported formatting wrappers, and matches a single A--D choice letter to the reference label for that ordering, ignoring letter case and allowing surrounding parentheses.

Extraction failures are counted as incorrect. Responses reaching the output limit are scored on the available final answer, and the session continues to the next problem. Responses that exhaust the model's context capacity count as incorrect. Process failures without a saved response remain missing. We report a turn-level score only when all four responses are present for every benchmark problem, and a T2--T4 mean only when all three turns are complete. The per-turn denominators are 2,000 responses for MATH-500, 120 for AIME 2025, 136 for AMC23$^{\dagger}$, and 792 for GPQA-Diamond.

When AMC23 was added to evaluation, normalized-text matching identified six problems present in all three training pools: original IDs $3,15,16,19,30,32$ in the test split of \texttt{math-ai/amc23}. These problems remain in training and are excluded from both current questions and session histories for every evaluated condition. The resulting 34-problem subset is marked with $\dagger$ in tables and figures.

\subsection{Shared-history Controls and Runtime Measurements}\label{app:component-runtime}

The T2 study uses Qwen3.5-4B and all 30 AIME 2025 problems, with four samples per problem. Its T1 trajectories were generated separately from those used in the continuous evaluation. Each of the 120 inputs contains a fixed T1 trajectory from the Native path. Exact-token replay through the frozen backbone constructs its historical bank using the capture boundaries in Appendix~\ref{app:method-capture}. All conditions use the same visible T1 history, current problem, generation seed, and decoding budget. Only T2 answers are evaluated in this study.

Native and the Bank-free controller use ordinary conversation history without reading the auxiliary bank. STAIR uses its final controller and the captured T1 bank. The Random-reflector control replaces its normals with fixed random unit directions under three seeds. The Direct reflected-read control replaces $r^R-r^{\mathrm{ref}}$ with $r^R$ while retaining the learned normals. Both modify the trained method at evaluation time.

The test-only K--V pairing control starts from the same captured bank and permutes values across positions within each K/V head. Keys and auxiliary positions remain fixed. Query heads sharing a K/V head use the same permutation, which remains fixed throughout an answer. Both K--V permutation conditions, Random-reflector, and the crossed random--permutation control each average three test-time configurations (360 responses each); the other conditions in Table~\ref{tab:shared-t1} cover 120. All conditions use the scoring rules in Appendix~\ref{app:evaluation-scoring}.

The Bank-free controller is trained separately with the same data, objective, update budget, controlled layers, and prefill positions as STAIR. Its 12,288 parameters are per-head, per-channel scales initialized to zero. During current-turn prefill, each scale multiplies the corresponding channel of the native attention output immediately before the frozen output projection. In Qwen3.5, the native output gate has already acted at this point. The controller reads no auxiliary bank.

The crossed comparison combines each saved random reflector with its matched value permutation. Current inputs, generation seeds, and bank coordinates remain fixed. Table~\ref{tab:reflector-pairing} reports all four cells: learned directions gain 11.11 Avg@4 points over random directions with intact pairing and lose 3.89 points with permuted pairing. Let $A$ denote Avg@4 in percent. The resulting interaction is
\begin{equation}
\begin{aligned}
\mathcal I
&=(A_{\mathrm{learned,intact}}-A_{\mathrm{random,intact}})\\
&\quad -(A_{\mathrm{learned,permuted}}-A_{\mathrm{random,permuted}})
=15.00\ \mathrm{pp}.
\end{aligned}
\label{eq:reflector-pairing-interaction}
\end{equation}

\begin{table}[H]
\centering
\setlength{\tabcolsep}{9pt}
\caption{Reflector--pairing comparison on shared-T1 AIME 2025 with Qwen3.5-4B at T2 (\%). Three matched test-time configurations are averaged, reusing the same 120 learned/intact responses in each comparison. Bold marks the maximum for each metric.}
\label{tab:reflector-pairing}
\begin{tabular}{lrrrr}
\toprule
& \multicolumn{2}{c}{Intact pairing} & \multicolumn{2}{c}{Permuted pairing}\\
\cmidrule(lr){2-3}\cmidrule(lr){4-5}
Reflector & Avg@4 & Pass@4 & Avg@4 & Pass@4\\
\midrule
Learned & \cellcolor{stairrow}\textbf{48.33} & \cellcolor{stairrow}\textbf{80.00} & 33.06 & 61.11\\
Random & 37.22 & 63.33 & 36.94 & 63.33\\
\bottomrule
\end{tabular}
\end{table}

To learn the readout under permuted K--V correspondence, we train a separate controller from initialization with the same data, CE plus Native KL objective, and 1,941-update budget as STAIR. The training sample identity and turn determine a value permutation for each K/V head. This permutation remains fixed within the read and its backward pass. Keys, positions, and bank length remain unchanged. Evaluation uses the same Shared-T1 T2 inputs and averages three test-time permutation seeds. This controller reaches 31.67\% Avg@4 and 58.89\% Pass@4 (Table~\ref{tab:shared-t1}); the three seeds describe evaluation interventions on this one trained controller.

Table~\ref{tab:shared-t1-intervals} reports paired differences and intervals for these shared-history controls, including the four-cell interaction.

The bank-prefix study uses the same 120 matched Qwen3.5-4B AIME 2025 inputs as the shared-T1 controls. It limits the readable bank to its first 2,048, 8,192, or 32,768 K/V entries in historical order; Full reads all entries. The visible conversation, stored K/V positions, and current-query virtual positions remain unchanged. A bank shorter than the budget is read in full. The median bank contains about 63.4K tokens. Table~\ref{tab:bank-prefix-accuracy} reports results for all 30 problems under every condition.

\begin{table}[H]
\centering
\setlength{\tabcolsep}{8pt}
\caption{Bank-prefix accuracy on shared-T1 AIME 2025 at T2 (\%). Each condition uses four matched sample seeds per problem. Truncated counts banks longer than the access budget.}
\label{tab:bank-prefix-accuracy}
\begin{tabular}{@{}lrrr@{}}
\toprule
Readable bank & Truncated & Avg@4 & Pass@4\\
\midrule
Native & -- & 21.67 & 56.67\\
\rowcolor{stairrow}STAIR 2K & 120/120 & 35.00 & 73.33\\
\rowcolor{stairrow}STAIR 8K & 114/120 & 43.33 & \textbf{80.00}\\
\rowcolor{stairrow}STAIR 32K & 81/120 & 39.17 & 70.00\\
\rowcolor{stairrow}STAIR Full & 0/120 & \textbf{48.33} & \textbf{80.00}\\
\bottomrule
\end{tabular}
\end{table}

\begin{samepage}
Fixed-workload runtime is measured on a single A800 80GB with Transformers, FlashAttention2, BF16, and batch size one. The 16 fixed Native MATH-500 T1 histories, selected without reference to answer correctness, each have a bank longer than 32K tokens and receive one warmup and three timed repetitions. We aggregate repetitions within each history, then report medians across histories (Table~\ref{tab:bank-runtime}). The Full bank contains a median 80.7K tokens. Relative to Native, its paired median increases are 567 ms for T2 prefill, 0.81 s for the fixed workload, and 6.60 GiB for peak allocated memory. Exact-token replay to construct the bank takes a further median 7.29 s and is excluded from this fixed-workload table.

\end{samepage}

\begin{table}[H]
\centering
\setlength{\tabcolsep}{4pt}
\renewcommand{\arraystretch}{1.1}
\caption{Bank-size and fixed-workload costs across 16 histories. Transfer moves the bank to the GPU; total includes transfer, current-question prefill, and a fixed 256-token decode. Entries summarize each history before taking the median across histories.}
\label{tab:bank-runtime}
\begin{tabular}{lrrrrrr}
\toprule
Bank & K/V & Transfer & Prefill & Decode & Peak & Total\\
 & (MiB) & (ms) & (ms) & (s) & (GiB) & (s)\\
\midrule
Native & 0 & $\approx 0$ & 70 & 10.08 & 8.03 & 10.15\\
2K & 24 & 6 & 110 & 10.02 & 8.19 & 10.15\\
8K & 96 & 24 & 157 & 10.06 & 8.70 & 10.23\\
32K & 384 & 138 & 338 & 10.05 & 10.71 & 10.56\\
\rowcolor{stairrow}Full & $\approx 946$ & 330 & 628 & 9.98 & 14.61 & 10.95\\
\bottomrule
\end{tabular}
\end{table}

To measure complete-session cost, we run four preselected AIME 2025 problem sequences with Qwen3.5-4B on an exclusive A100 GPU, using Transformers, FlashAttention2, BF16, and batch size one. Each method generates all four answers in every sequence, giving 48 generations. Vanilla resets the context for each problem. Native and STAIR continue from their own generated histories, and STAIR captures its actual earlier K/V states. All methods re-prefill their ordinary prompts at each turn. The timer includes prompt processing, answer generation, and bank capture, storage, and transfer. Shared model loading is excluded, and method order rotates across sequences.

\begin{table}[H]
\centering
\setlength{\tabcolsep}{5pt}
\caption{Complete four-turn session cost on AIME 2025 with Qwen3.5-4B and an exclusive A100 GPU. Time, output tokens, and prefill are means over the same four sequences. Memory and auxiliary-bank columns report the mean / maximum of session peaks. Output counts include reasoning and final answers.}
\label{tab:session-runtime}
\begin{tabular}{lrrrrr}
\toprule
Condition & Time & Output tokens & Prefill & Peak memory & Auxiliary bank\\
& (min) & & (s) & (GiB) & (GiB)\\
\midrule
Vanilla & 107.93 & 196,753 & 0.777 & 10.68 / 10.69 & 0 / 0\\
Native & 98.25 & 178,216 & 0.717 & 10.54 / 10.81 & 0 / 0\\
\rowcolor{stairrow}STAIR & 116.00 & 208,877 & 1.442 & 17.24 / 19.70 & 1.66 / 2.15\\
\bottomrule
\end{tabular}
\end{table}

\begin{samepage}
In these complete sessions, STAIR takes 18.1\% longer than Native and generates 17.2\% more tokens. These times describe the sampled answer trajectories, including their different output lengths. We sum prefill time over the four turns and take the largest allocated-memory and auxiliary-bank sizes within each session. Bank sizes count the separately captured historical K/V entries. The fixed-workload measurement in Table~\ref{tab:bank-runtime} holds the decode length constant.

\end{samepage}

\subsection{History-management Comparison}\label{app:history-management}

This comparison uses 120 saved Native T1--T3 histories for Qwen3.5-4B on AIME 2025. Every condition generates new T4 responses with matched current problems, sample seeds, and decoding budgets. All responses use the same final-answer scoring rule. Ordinary visible history contains earlier questions and the assistant answers after \texttt{</think>}. STAIR also accesses historical reasoning K/V states captured by exact-token replay of those earlier turns.

We evaluate all ten conditions under three groups of decoding seeds, retaining the same histories, prepared summaries, selection plans, and controller checkpoints. Each group contributes 120 responses per condition. We compute Avg@4 and Pass@4 from four responses per problem within each group, then report the arithmetic mean and sample standard deviation across groups (Table~\ref{tab:history-management}). Table~\ref{tab:history-management-seeds} gives the group-level scores.

History summarization uses frozen Qwen3.5-4B with thinking disabled and greedy decoding. It compresses the three earlier questions and their visible answers into at most 512 tokens, without access to the current question. The summary replaces the earlier turns. STAIR + Summarization reuses the same saved summaries and ordinary prompts. Its unchanged controller reads the original Native T1--T3 reasoning bank, retaining its positions and full-bank query offset. Turn selection uses the same frozen model and decoding mode. Given the visible history and current question, decoding is constrained to T1, T2, T3, or NONE. We retain only the selected question and its original visible answer; NONE removes all earlier turns.

Sparse access keeps the input token sequence unchanged. An additional Native prefill scores historical keys with current-turn queries. Within each full-attention layer, we apply softmax to query--key scores over eligible history positions only. We average the resulting weights over current-turn prompt positions and query heads sharing each K/V head, then retain the top 25\% or 50\% of historical content positions per K/V head. Role boundaries, current-turn positions, and generated tokens remain accessible. The selection stays fixed throughout the answer, and positions retain their original encodings. Gated DeltaNet layers are unchanged. For STAIR combined with sparse access, we separately select the same fraction of auxiliary-bank entries. Reference and reflected reads share this selection and retain the original positions. The percentages specify historical access budgets.

\begin{table}[H]
\centering
\setlength{\tabcolsep}{9pt}
\renewcommand{\arraystretch}{1.12}
\caption{Mean token counts per response in the original T4 evaluation group. Prompt and response count tokens used to answer the current problem, including reasoning. Auxiliary output counts summary text or a turn-selection label.}
\label{tab:history-management-tokens}
\begin{tabular}{lrrr}
\toprule
Condition & Prompt & Response & Auxiliary output\\
\midrule
Native & 3,353 & 53,498 & --\\
History summarization & 717 & 47,474 & 506\\
Turn selection & 1,471 & 65,853 & $\approx 2$\\
Sparse-25\% & 3,353 & 53,066 & --\\
Sparse-50\% & 3,353 & 52,353 & --\\
\rowcolor{stairrow}STAIR & 3,353 & 53,402 & --\\
\bottomrule
\end{tabular}
\end{table}

The summarization and turn-selection calls receive an average of 3,215 and 3,417 input tokens. Their batch-amortized preparation times are 3.11 s and 0.95 s per response, excluding answer generation. Preparing cached selection plans for both sparse budgets, including auxiliary-bank selections, together takes a mean 0.447 s per input. This timing excludes bank construction and answer generation.

The reference-read gate uses the same full bank and ordinary prompt as STAIR. It computes $r^{\mathrm{ref}}$ with the unreflected query, then multiplies each head's readout channels by learned scales before the native output gate and frozen output projection. The 12,288 scales are initialized to zero at layers $[3,11,19]$. Training matches STAIR's data, T2--T4 supervision, CE plus Native KL objective, and 1,941-update budget, using AdamW with a constant learning rate of 0.003 and zero weight decay. Evaluation uses the final checkpoint and the shared-history T4 inputs.

\subsection{Llama-backbone Extension}\label{app:llama-extension}

The backbone is \texttt{deepseek-ai/DeepSeek-R1-Distill-Llama-8B} \citep{deepseek2025r1llama8b}. We train its controller on the published OpenThoughts 32K response collection for this model \citep{singhi2025openthoughtsllama}. The collection contains 22,037 responses covering 11,072 normalized problem identities. A problem-level split assigns 10,944 problems to training and 128 to validation. Normalized-text matching found no AIME 2025 overlaps. Responses are retained irrespective of correctness.

All 21,781 training responses serve once as supervised targets. Three distinct target problems form T2--T4; a saved response to a different training problem supplies history-only T1. Problem-balancing weights give each training problem equal total weight. Training uses the same CE plus Native KL objective, AdamW settings, and eight-session update batch as the Qwen runs. We train 12,288 reflector parameters at layers $[3,11,19]$ for one pass, totaling 908 updates, and use the final checkpoint. The auxiliary read uses the backbone's native Llama Q/K projections, scaled rotary embedding, and output projection. Its control span runs from the current user-role header through the assistant thinking prefix.

Evaluation uses all 30 AIME 2025 problems with four samples per problem in four-turn sessions. Generation uses BF16, Transformers, and FlashAttention2, with temperature 0.6, top-$p$ 0.95, unrestricted top-$k$, min-$p$ 0, presence penalty 0, repetition penalty 1, and a 32,768-token output limit. The official chat template supplies the thinking prefix, with instructions in the user message and no additional system prompt. Native and STAIR share the no-history T1 responses and use matched problem schedules and sample seeds. Each continues from its own later generated history. Visible history retains earlier problems and post-thinking answers; STAIR captures reasoning states in its auxiliary bank. Scoring follows Appendix~\ref{app:evaluation-scoring}.

We repeat this evaluation under three decoding-seed groups with the same final controller. Each group computes Avg@4 and Pass@4 from four samples per problem at each turn, then averages T2--T4. Table~\ref{tab:llama-evaluation-seeds} reports these group-level means and their mean and sample standard deviation across groups. Table~\ref{tab:turns-r1_distill_llama8b} gives the original group's per-turn results.

\subsection{Transfer to Code Generation}\label{app:code-transfer}

We evaluate Qwen3.5-9B on all 164 HumanEval+ tasks \citep{liu2023evalplus}, using its original math-trained controller without code-domain fine-tuning. The four-turn protocol follows Appendix~\ref{app:evaluation-sampling}, with four samples per task and each condition continuing from its own generated histories. Native and STAIR share the no-history T1 responses, which also provide the Vanilla reference. Functional correctness is scored by executing the generated code with the official EvalPlus test suite. Avg@4 measures the fraction of correct samples, and Pass@4 measures the fraction of tasks with at least one correct sample. Table~\ref{tab:turns-humanevalplus} reports every turn and the T2--T4 means; Table~\ref{tab:extension-intervals} gives paired mean differences and intervals.

\subsection{Training-stage Comparison}\label{app:stage-protocol}

We compare DeepSeek-R1-Distill-Qwen-1.5B \citep{deepseek2025r1qwen15b}, a model trained by supervised reasoning distillation, with DeepScaleR-1.5B-Preview \citep{agentica2025deepscaler}, its reinforcement-learning continuation. Each checkpoint answers all 30 AIME 2025 problems with four samples under Vanilla, Native, and STAIR. Native and STAIR are evaluated at T2 after the same SFT-generated T1 history texts. Each checkpoint replays those texts to construct its own K/V states. Current-question and sample seeds are matched across stages and methods. Sampling uses temperature 0.6, top-$p$ 0.95, top-$k$ disabled, repetition penalty one, no presence penalty, and a 32,768-token output limit. Both stages use a thinking prefix and no added system message.

Both controllers train on the same Qwen3.5-4B DAPO response texts, problem splits, and four-turn session assignments. Each model's tokenizer encodes these texts independently, and each backbone produces its own historical states and Native teacher distribution. The objective, AdamW settings, and eight-session update batch follow Appendix~\ref{app:training-settings}. Each controller trains 4,608 reflector parameters at layers $[3,11,19]$ for 1,941 updates; evaluation uses the final checkpoint.

We evaluate all six stage--method conditions under three decoding-seed groups, with the shared T1 texts and trained controllers fixed. Each condition produces 120 responses per group. Scoring follows Appendix~\ref{app:evaluation-scoring}; Avg@4 and Pass@4 are computed within each group using four responses per problem. Table~\ref{tab:training-stages} reports means and sample standard deviations across groups, and Table~\ref{tab:stage-seeds} gives every group's scores.

\clearpage
\section{Continuous Reasoning Results}\label{app:continuous-results}
Tables~\ref{tab:turns-qwen3_4b_instruct}--\ref{tab:turns-humanevalplus} report Avg@4 and Pass@4 in percent by turn and as T2--T4 means. Bold marks the highest continuous-condition score at each later turn and in the mean, including ties. Dashes indicate turns not applicable to Vanilla.
\subsection{Qwen3-4B Instruct}
\begin{table}[H]
\centering
\setlength{\tabcolsep}{2pt}
\caption{Qwen3-4B Instruct: results by turn (\%). Mean averages T2--T4.}
\label{tab:turns-qwen3_4b_instruct}
\begin{tabular}{@{}l*{10}{r}@{}}
\toprule
& \multicolumn{2}{c}{T1} & \multicolumn{2}{c}{T2} & \multicolumn{2}{c}{T3} & \multicolumn{2}{c}{T4} & \multicolumn{2}{c}{Mean}\\
\cmidrule(lr){2-3}\cmidrule(lr){4-5}\cmidrule(lr){6-7}\cmidrule(lr){8-9}\cmidrule(lr){10-11}
Condition & Avg@4 & Pass@4 & Avg@4 & Pass@4 & Avg@4 & Pass@4 & Avg@4 & Pass@4 & Avg@4 & Pass@4\\
\midrule
\multicolumn{11}{@{}l}{\textbf{\textit{MATH-500}}}\\
Vanilla & 95.00 & 97.20 & -- & -- & -- & -- & -- & -- & -- & --\\
Native & 95.00 & 97.20 & 94.50 & 97.40 & 94.05 & \textbf{97.60} & 93.75 & \textbf{97.80} & 94.10 & \textbf{97.60}\\
\rowcolor{stairrow}\textbf{STAIR} & 95.00 & 97.20 & \textbf{95.00} & \textbf{97.60} & \textbf{94.95} & \textbf{97.60} & \textbf{93.95} & 97.40 & \textbf{94.63} & 97.53\\
\addlinespace
\multicolumn{11}{@{}l}{\textbf{\textit{AIME 2025}}}\\
Vanilla & 47.50 & 66.67 & -- & -- & -- & -- & -- & -- & -- & --\\
Native & 47.50 & 66.67 & 36.67 & 50.00 & 32.50 & 40.00 & 34.17 & \textbf{53.33} & 34.44 & 47.78\\
\rowcolor{stairrow}\textbf{STAIR} & 47.50 & 66.67 & \textbf{40.00} & \textbf{53.33} & \textbf{35.83} & \textbf{50.00} & \textbf{38.33} & \textbf{53.33} & \textbf{38.06} & \textbf{52.22}\\
\addlinespace
\multicolumn{11}{@{}l}{\textbf{\textit{AMC23$^{\dagger}$}}}\\
Vanilla & 92.65 & 97.06 & -- & -- & -- & -- & -- & -- & -- & --\\
Native & 92.65 & 97.06 & 94.12 & 97.06 & \textbf{93.38} & \textbf{97.06} & 92.65 & 94.12 & \textbf{93.38} & 96.08\\
\rowcolor{stairrow}\textbf{STAIR} & 92.65 & 97.06 & \textbf{94.85} & \textbf{100.00} & 91.18 & 94.12 & \textbf{94.12} & \textbf{97.06} & \textbf{93.38} & \textbf{97.06}\\
\addlinespace
\multicolumn{11}{@{}l}{\textbf{\textit{GPQA-Diamond}}}\\
Vanilla & 58.59 & 78.79 & -- & -- & -- & -- & -- & -- & -- & --\\
Native & 58.59 & 78.79 & \textbf{60.86} & \textbf{78.28} & 56.19 & 74.75 & 56.94 & \textbf{77.78} & 58.00 & 76.94\\
\rowcolor{stairrow}\textbf{STAIR} & 58.59 & 78.79 & 58.46 & 75.76 & \textbf{58.84} & \textbf{80.30} & \textbf{58.96} & 77.27 & \textbf{58.75} & \textbf{77.78}\\
\addlinespace
\bottomrule
\end{tabular}
\par\smallskip\noindent$\dagger$ AMC23 excludes problems overlapping the training set, leaving 34 problems.
\end{table}
\subsection{LoRA Comparison and Joint Training on AIME 2025}
\begin{table}[H]
\centering
\setlength{\tabcolsep}{2pt}
\caption{Qwen3-4B Instruct on AIME 2025: standalone and jointly trained adapters by turn (\%). Mean averages T2--T4; bold marks the highest score in each column, including ties.}
\label{tab:lora-turns}
\begin{tabular}{@{}l*{8}{r}@{}}
\toprule
& \multicolumn{2}{c}{T2} & \multicolumn{2}{c}{T3} & \multicolumn{2}{c}{T4} & \multicolumn{2}{c}{Mean}\\
\cmidrule(lr){2-3}\cmidrule(lr){4-5}\cmidrule(lr){6-7}\cmidrule(lr){8-9}
Condition & Avg@4 & Pass@4 & Avg@4 & Pass@4 & Avg@4 & Pass@4 & Avg@4 & Pass@4\\
\midrule
Native & 36.67 & 50.00 & 32.50 & 40.00 & 34.17 & 53.33 & 34.44 & 47.78\\
\rowcolor{stairrow}\textbf{STAIR} & 40.00 & 53.33 & 35.83 & 50.00 & 38.33 & 53.33 & 38.06 & 52.22\\
LoRA & 40.83 & 53.33 & 38.33 & 60.00 & 39.17 & \textbf{63.33} & 39.44 & 58.89\\
\rowcolor{stairrow}\textbf{LoRA + STAIR} & \textbf{44.17} & \textbf{63.33} & \textbf{45.00} & \textbf{70.00} & \textbf{40.83} & 56.67 & \textbf{43.33} & \textbf{63.33}\\
\bottomrule
\end{tabular}
\end{table}

\vfill\clearpage
\subsection{Qwen3.5-4B}
\begin{table}[H]
\centering
\setlength{\tabcolsep}{2pt}
\caption{Qwen3.5-4B: results by turn (\%). Mean averages T2--T4.}
\label{tab:turns-qwen35_4b_thinking}
\begin{tabular}{@{}l*{10}{r}@{}}
\toprule
& \multicolumn{2}{c}{T1} & \multicolumn{2}{c}{T2} & \multicolumn{2}{c}{T3} & \multicolumn{2}{c}{T4} & \multicolumn{2}{c}{Mean}\\
\cmidrule(lr){2-3}\cmidrule(lr){4-5}\cmidrule(lr){6-7}\cmidrule(lr){8-9}\cmidrule(lr){10-11}
Condition & Avg@4 & Pass@4 & Avg@4 & Pass@4 & Avg@4 & Pass@4 & Avg@4 & Pass@4 & Avg@4 & Pass@4\\
\midrule
\multicolumn{11}{@{}l}{\textbf{\textit{MATH-500}}}\\
Vanilla & 96.00 & 98.60 & -- & -- & -- & -- & -- & -- & -- & --\\
Native & 96.00 & 98.60 & 90.50 & 98.20 & 96.55 & 99.20 & 97.30 & \textbf{99.00} & 94.78 & 98.80\\
\rowcolor{stairrow}\textbf{STAIR} & 96.00 & 98.60 & \textbf{93.60} & \textbf{99.40} & \textbf{97.45} & \textbf{99.40} & \textbf{97.75} & \textbf{99.00} & \textbf{96.27} & \textbf{99.27}\\
\addlinespace
\multicolumn{11}{@{}l}{\textbf{\textit{AIME 2025}}}\\
Vanilla & 60.83 & 80.00 & -- & -- & -- & -- & -- & -- & -- & --\\
Native & 60.83 & 80.00 & 30.00 & 53.33 & 44.17 & 63.33 & 47.50 & 76.67 & 40.56 & 64.44\\
\rowcolor{stairrow}\textbf{STAIR} & 60.83 & 80.00 & \textbf{42.50} & \textbf{76.67} & \textbf{50.00} & \textbf{76.67} & \textbf{64.17} & \textbf{80.00} & \textbf{52.22} & \textbf{77.78}\\
\addlinespace
\multicolumn{11}{@{}l}{\textbf{\textit{AMC23$^{\dagger}$}}}\\
Vanilla & 76.47 & 85.29 & -- & -- & -- & -- & -- & -- & -- & --\\
Native & 76.47 & 85.29 & 52.21 & 73.53 & 73.53 & \textbf{97.06} & \textbf{78.68} & \textbf{91.18} & 68.14 & 87.25\\
\rowcolor{stairrow}\textbf{STAIR} & 76.47 & 85.29 & \textbf{60.29} & \textbf{82.35} & \textbf{84.56} & \textbf{97.06} & \textbf{78.68} & \textbf{91.18} & \textbf{74.51} & \textbf{90.20}\\
\addlinespace
\multicolumn{11}{@{}l}{\textbf{\textit{GPQA-Diamond}}}\\
Vanilla & 63.64 & 75.76 & -- & -- & -- & -- & -- & -- & -- & --\\
Native & 63.64 & 75.76 & 75.25 & \textbf{89.39} & \textbf{75.25} & \textbf{88.89} & \textbf{75.25} & \textbf{89.90} & 75.25 & \textbf{89.39}\\
\rowcolor{stairrow}\textbf{STAIR} & 63.64 & 75.76 & \textbf{77.53} & \textbf{89.39} & 74.75 & 86.87 & 73.86 & 84.34 & \textbf{75.38} & 86.87\\
\addlinespace
\bottomrule
\end{tabular}
\par\smallskip\noindent$\dagger$ AMC23 excludes problems overlapping the training set, leaving 34 problems.
\end{table}
\subsection{Qwen3.5-9B}
\begin{table}[H]
\centering
\setlength{\tabcolsep}{2pt}
\caption{Qwen3.5-9B: results by turn (\%). Mean averages T2--T4.}
\label{tab:turns-qwen35_9b_thinking}
\begin{tabular}{@{}l*{10}{r}@{}}
\toprule
& \multicolumn{2}{c}{T1} & \multicolumn{2}{c}{T2} & \multicolumn{2}{c}{T3} & \multicolumn{2}{c}{T4} & \multicolumn{2}{c}{Mean}\\
\cmidrule(lr){2-3}\cmidrule(lr){4-5}\cmidrule(lr){6-7}\cmidrule(lr){8-9}\cmidrule(lr){10-11}
Condition & Avg@4 & Pass@4 & Avg@4 & Pass@4 & Avg@4 & Pass@4 & Avg@4 & Pass@4 & Avg@4 & Pass@4\\
\midrule
\multicolumn{11}{@{}l}{\textbf{\textit{MATH-500}}}\\
Vanilla & 96.65 & 98.20 & -- & -- & -- & -- & -- & -- & -- & --\\
Native & 96.65 & 98.20 & 94.30 & \textbf{98.80} & \textbf{97.30} & \textbf{99.60} & 97.55 & \textbf{99.20} & 96.38 & \textbf{99.20}\\
\rowcolor{stairrow}\textbf{STAIR} & 96.65 & 98.20 & \textbf{95.15} & \textbf{98.80} & 97.10 & 99.40 & \textbf{98.10} & \textbf{99.20} & \textbf{96.78} & 99.13\\
\addlinespace
\multicolumn{11}{@{}l}{\textbf{\textit{AIME 2025}}}\\
Vanilla & 61.67 & 73.33 & -- & -- & -- & -- & -- & -- & -- & --\\
Native & 61.67 & 73.33 & 47.50 & 70.00 & 39.17 & 66.67 & 43.33 & 70.00 & 43.33 & 68.89\\
\rowcolor{stairrow}\textbf{STAIR} & 61.67 & 73.33 & \textbf{51.67} & \textbf{73.33} & \textbf{47.50} & \textbf{70.00} & \textbf{49.17} & \textbf{76.67} & \textbf{49.44} & \textbf{73.33}\\
\addlinespace
\multicolumn{11}{@{}l}{\textbf{\textit{AMC23$^{\dagger}$}}}\\
Vanilla & 77.94 & 85.29 & -- & -- & -- & -- & -- & -- & -- & --\\
Native & 77.94 & 85.29 & 66.91 & \textbf{91.18} & 72.79 & 91.18 & \textbf{77.94} & 91.18 & 72.55 & 91.18\\
\rowcolor{stairrow}\textbf{STAIR} & 77.94 & 85.29 & \textbf{71.32} & 85.29 & \textbf{80.15} & \textbf{94.12} & 75.74 & \textbf{97.06} & \textbf{75.74} & \textbf{92.16}\\
\addlinespace
\multicolumn{11}{@{}l}{\textbf{\textit{GPQA-Diamond}}}\\
Vanilla & 68.81 & 80.30 & -- & -- & -- & -- & -- & -- & -- & --\\
Native & 68.81 & 80.30 & \textbf{79.80} & \textbf{90.91} & \textbf{79.92} & \textbf{89.39} & \textbf{79.92} & 88.89 & \textbf{79.88} & \textbf{89.73}\\
\rowcolor{stairrow}\textbf{STAIR} & 68.81 & 80.30 & 72.10 & 85.86 & 76.89 & 88.38 & 78.79 & \textbf{91.41} & 75.93 & 88.55\\
\addlinespace
\bottomrule
\end{tabular}
\par\smallskip\noindent$\dagger$ AMC23 excludes problems overlapping the training set, leaving 34 problems.
\end{table}

\subsection{DeepSeek-R1-Distill-Llama-8B}
\begin{table}[H]
\centering
\setlength{\tabcolsep}{2pt}
\caption{DeepSeek-R1-Distill-Llama-8B on AIME 2025: results by turn in the original evaluation group (\%). Mean averages T2--T4. Native and STAIR share the no-history T1 reference.}
\label{tab:turns-r1_distill_llama8b}
\begin{tabular}{@{}l*{10}{r}@{}}
\toprule
& \multicolumn{2}{c}{T1} & \multicolumn{2}{c}{T2} & \multicolumn{2}{c}{T3} & \multicolumn{2}{c}{T4} & \multicolumn{2}{c}{Mean}\\
\cmidrule(lr){2-3}\cmidrule(lr){4-5}\cmidrule(lr){6-7}\cmidrule(lr){8-9}\cmidrule(lr){10-11}
Condition & Avg@4 & Pass@4 & Avg@4 & Pass@4 & Avg@4 & Pass@4 & Avg@4 & Pass@4 & Avg@4 & Pass@4\\
\midrule
Vanilla & 33.33 & 46.67 & -- & -- & -- & -- & -- & -- & -- & --\\
Native & 33.33 & 46.67 & 20.00 & 33.33 & \textbf{30.00} & \textbf{46.67} & 22.50 & \textbf{40.00} & 24.17 & 40.00\\
\rowcolor{stairrow}\textbf{STAIR} & 33.33 & 46.67 & \textbf{33.33} & \textbf{46.67} & 25.00 & 40.00 & \textbf{26.67} & 36.67 & \textbf{28.33} & \textbf{41.11}\\
\bottomrule
\end{tabular}
\end{table}

\begin{table}[H]
\centering
\setlength{\tabcolsep}{8pt}
\renewcommand{\arraystretch}{1.12}
\caption{DeepSeek-R1-Distill-Llama-8B on AIME 2025 across three decoding-seed groups (\%). Group columns report Avg@4 / Pass@4, averaged over T2--T4. Mean columns give the three-group mean with sample standard deviation in subscript. Bold compares Native and STAIR.}
\label{tab:llama-evaluation-seeds}
\begin{tabular}{lrrrrr}
\toprule
& & & & \multicolumn{2}{c}{Mean}\\
\cmidrule(lr){5-6}
Condition & Original & Seed 1 & Seed 2 & Avg@4 & Pass@4\\
\midrule
Vanilla & 33.33 / 46.67 & 35.00 / 50.00 & 34.17 / 56.67 & \meanstd{34.17}{0.83} & \meanstd{51.11}{5.09}\\
Native & 24.17 / 40.00 & 23.61 / 41.11 & 24.17 / 44.44 & \meanstd{23.98}{0.32} & \meanstd{41.85}{2.31}\\
\rowcolor{stairrow}\textbf{STAIR} & \textbf{28.33} / \textbf{41.11} & \textbf{28.61} / \textbf{44.44} & \textbf{29.44} / \textbf{45.56} & \meanstd{\textbf{28.80}}{0.58} & \meanstd{\textbf{43.70}}{2.31}\\
\bottomrule
\end{tabular}
\end{table}

\subsection{HumanEval+ Code Generation}\label{app:code-results}
\begin{table}[H]
\centering
\setlength{\tabcolsep}{2pt}
\caption{Qwen3.5-9B on HumanEval+ (\%), using its math-trained controller without further training. Mean averages T2--T4. Native and STAIR share T1; scoring uses the official EvalPlus tests.}
\label{tab:turns-humanevalplus}
\begin{tabular}{@{}l*{10}{r}@{}}
\toprule
& \multicolumn{2}{c}{T1} & \multicolumn{2}{c}{T2} & \multicolumn{2}{c}{T3} & \multicolumn{2}{c}{T4} & \multicolumn{2}{c}{Mean}\\
\cmidrule(lr){2-3}\cmidrule(lr){4-5}\cmidrule(lr){6-7}\cmidrule(lr){8-9}\cmidrule(lr){10-11}
Condition & Avg@4 & Pass@4 & Avg@4 & Pass@4 & Avg@4 & Pass@4 & Avg@4 & Pass@4 & Avg@4 & Pass@4\\
\midrule
Vanilla & 87.50 & 92.68 & -- & -- & -- & -- & -- & -- & -- & --\\
Native & 87.50 & 92.68 & 85.21 & 93.29 & 85.52 & 93.90 & 87.80 & \textbf{94.51} & 86.18 & 93.90\\
\rowcolor{stairrow}\textbf{STAIR} & 87.50 & 92.68 & \textbf{86.74} & \textbf{93.90} & \textbf{88.72} & \textbf{94.51} & \textbf{88.26} & \textbf{94.51} & \textbf{87.91} & \textbf{94.31}\\
\bottomrule
\end{tabular}
\end{table}

\clearpage
\subsection{Training-stage Comparison}\label{app:stage-results}

STAIR improves mean Avg@4 and Pass@4 over Native at both training stages (Table~\ref{tab:training-stages}). Its Avg@4 gains average 5.28 points at the SFT stage and 6.39 at the RL stage, with positive differences in all three seed groups at each stage. Table~\ref{tab:stage-seeds} reports the group-level scores under the shared-history protocol in Appendix~\ref{app:stage-protocol}.

\begin{table}[H]
\centering
\setlength{\tabcolsep}{7pt}
\renewcommand{\arraystretch}{1.12}
\caption{AIME 2025 results at two stages of the same 1.5B training lineage (\%). Entries are means over three decoding-seed groups, with sample standard deviations in subscripts. Native and STAIR share SFT-generated T1 texts and are evaluated at T2 using the final stage-specific controllers. Bold compares Native and STAIR within each stage.}
\label{tab:training-stages}
\begin{tabular}{llrr}
\toprule
Stage & Condition & Avg@4 & Pass@4\\
\midrule
SFT (R1-Distill) & Vanilla & \meanstd{25.56}{0.48} & \meanstd{37.78}{5.09}\\
& Native & \meanstd{14.72}{2.68} & \meanstd{26.67}{3.33}\\
\rowcolor{stairrow}& \textbf{STAIR} & \meanstd{\textbf{20.00}}{1.67} & \meanstd{\textbf{32.22}}{1.92}\\
\midrule
RL (DeepScaleR) & Vanilla & \meanstd{28.89}{1.92} & \meanstd{37.78}{3.85}\\
& Native & \meanstd{20.83}{1.44} & \meanstd{32.22}{6.94}\\
\rowcolor{stairrow}& \textbf{STAIR} & \meanstd{\textbf{27.22}}{1.73} & \meanstd{\textbf{36.67}}{3.33}\\
\bottomrule
\end{tabular}
\end{table}

\begin{table}[H]
\centering
\setlength{\tabcolsep}{9pt}
\renewcommand{\arraystretch}{1.12}
\caption{Group-level AIME 2025 results for the 1.5B training-stage comparison. Each cell reports Avg@4 / Pass@4 (\%) from 120 responses. Bold compares Native and STAIR within each stage and seed group.}
\label{tab:stage-seeds}
\begin{tabular}{llrrr}
\toprule
Stage & Condition & Original & Seed 1 & Seed 2\\
\midrule
SFT & Vanilla & 25.83 / 33.33 & 25.00 / 43.33 & 25.83 / 36.67\\
& Native & 11.67 / 30.00 & 16.67 / 26.67 & 15.83 / 23.33\\
\rowcolor{stairrow}& \textbf{STAIR} & \textbf{21.67} / \textbf{33.33} & \textbf{20.00} / \textbf{33.33} & \textbf{18.33} / \textbf{30.00}\\
\midrule
RL & Vanilla & 26.67 / 33.33 & 30.00 / 40.00 & 30.00 / 40.00\\
& Native & 19.17 / 26.67 & 21.67 / \textbf{40.00} & 21.67 / 30.00\\
\rowcolor{stairrow}& \textbf{STAIR} & \textbf{25.83} / \textbf{33.33} & \textbf{26.67} / 36.67 & \textbf{29.17} / \textbf{40.00}\\
\bottomrule
\end{tabular}
\end{table}

\clearpage
\subsection{Repeated Evaluation and Paired Uncertainty}\label{app:paired-intervals}

For the main Qwen natural rollouts, we estimate uncertainty using a paired multiplier bootstrap with crossed question and actual-session clusters. Each of 4,000 replicates draws independent $\operatorname{Exp}(1)$ weights for each cluster dimension and multiplies the applicable question and session weights. The four samples stay together, and matched conditions receive the same weights. We recompute the paired score differences and use their 2.5th and 97.5th percentiles as exploratory pointwise 95\% intervals. Controllers remain fixed. These intervals describe benchmark-outcome uncertainty conditional on the fitted models.

Tables~\ref{tab:paired-qwen3_4b_instruct}--\ref{tab:paired-qwen35_9b_thinking} report all turns and T2--T4 means for both metrics. Differences are computed before rounding. Mean is the arithmetic mean of the three turn-level scores.

\begin{table}[H]
\centering
\setlength{\tabcolsep}{8pt}
\renewcommand{\arraystretch}{1.1}
\caption{Qwen3-4B Instruct: paired differences and pointwise 95\% intervals, in percentage points.}
\label{tab:paired-qwen3_4b_instruct}
\begin{tabular*}{\linewidth}{@{\extracolsep{\fill}}llrr@{}}
\toprule
Turn & Metric & STAIR $-$ Native & STAIR $-$ Vanilla\\
\midrule
\multicolumn{4}{@{}l}{\textbf{\textit{MATH-500}}}\\
T2 & Avg@4 & $+0.50\;[-1.11, +2.04]$ & $+0.00\;[-1.59, +1.81]$\\
T2 & Pass@4 & $+0.20\;[-1.28, +1.89]$ & $+0.40\;[-1.24, +2.23]$\\
T3 & Avg@4 & $+0.90\;[-0.72, +2.62]$ & $-0.05\;[-1.73, +1.67]$\\
T3 & Pass@4 & $+0.00\;[-2.10, +1.98]$ & $+0.40\;[-1.82, +2.80]$\\
T4 & Avg@4 & $+0.20\;[-1.27, +1.74]$ & $-1.05\;[-2.90, +0.77]$\\
T4 & Pass@4 & $-0.40\;[-2.25, +1.13]$ & $+0.20\;[-1.33, +1.85]$\\
Mean & Avg@4 & $+0.53\;[-0.41, +1.48]$ & $-0.37\;[-1.48, +0.90]$\\
Mean & Pass@4 & $-0.07\;[-0.97, +0.84]$ & $+0.33\;[-0.81, +1.66]$\\
\addlinespace
\multicolumn{4}{@{}l}{\textbf{\textit{AIME 2025}}}\\
T2 & Avg@4 & $+3.33\;[-9.69, +16.87]$ & $-7.50\;[-20.41, +1.87]$\\
T2 & Pass@4 & $+3.33\;[+0.02, +18.19]$ & $-13.33\;[-38.00, -1.81]$\\
T3 & Avg@4 & $+3.33\;[-7.10, +17.76]$ & $-11.67\;[-24.80, -2.16]$\\
T3 & Pass@4 & $+10.00\;[+0.79, +33.30]$ & $-16.67\;[-42.69, -2.73]$\\
T4 & Avg@4 & $+4.17\;[-6.71, +17.45]$ & $-9.17\;[-23.93, +1.71]$\\
T4 & Pass@4 & $+0.00\;[-26.54, +26.92]$ & $-13.33\;[-38.01, -1.70]$\\
Mean & Avg@4 & $+3.61\;[-2.93, +12.65]$ & $-9.44\;[-18.62, -1.98]$\\
Mean & Pass@4 & $+4.44\;[-6.57, +18.25]$ & $-14.44\;[-31.61, -3.74]$\\
\addlinespace
\multicolumn{4}{@{}l}{\textbf{\textit{AMC23$^{\dagger}$}}}\\
T2 & Avg@4 & $+0.74\;[-3.08, +5.52]$ & $+2.21\;[-2.29, +10.23]$\\
T2 & Pass@4 & $+2.94\;[+0.01, +16.77]$ & $+2.94\;[+0.01, +17.57]$\\
T3 & Avg@4 & $-2.21\;[-10.42, +3.86]$ & $-1.47\;[-8.13, +4.14]$\\
T3 & Pass@4 & $-2.94\;[-17.01, -0.01]$ & $-2.94\;[-17.01, -0.01]$\\
T4 & Avg@4 & $+1.47\;[-2.48, +6.94]$ & $+1.47\;[-2.46, +7.19]$\\
T4 & Pass@4 & $+2.94\;[+0.01, +15.22]$ & $+0.00\;[-13.41, +13.43]$\\
Mean & Avg@4 & $+0.00\;[-3.19, +3.03]$ & $+0.74\;[-2.44, +4.94]$\\
Mean & Pass@4 & $+0.98\;[-3.37, +6.71]$ & $+0.00\;[-7.75, +8.31]$\\
\addlinespace
\multicolumn{4}{@{}l}{\textbf{\textit{GPQA-Diamond}}}\\
T2 & Avg@4 & $-2.40\;[-8.23, +3.35]$ & $-0.13\;[-6.56, +6.28]$\\
T2 & Pass@4 & $-2.53\;[-10.40, +4.98]$ & $-3.03\;[-12.24, +5.31]$\\
T3 & Avg@4 & $+2.65\;[-2.63, +7.71]$ & $+0.25\;[-5.27, +5.45]$\\
T3 & Pass@4 & $+5.56\;[-2.20, +14.96]$ & $+1.52\;[-6.66, +10.17]$\\
T4 & Avg@4 & $+2.02\;[-3.09, +7.46]$ & $+0.38\;[-4.84, +5.48]$\\
T4 & Pass@4 & $-0.51\;[-8.30, +7.34]$ & $-1.52\;[-9.00, +5.65]$\\
Mean & Avg@4 & $+0.76\;[-2.53, +4.13]$ & $+0.17\;[-3.69, +4.03]$\\
Mean & Pass@4 & $+0.84\;[-3.57, +5.83]$ & $-1.01\;[-6.30, +4.32]$\\
\bottomrule
\end{tabular*}
\end{table}

\clearpage
\begin{table}[H]
\centering
\setlength{\tabcolsep}{8pt}
\renewcommand{\arraystretch}{1.1}
\caption{Qwen3.5-4B: paired differences and pointwise 95\% intervals, in percentage points.}
\label{tab:paired-qwen35_4b_thinking}
\begin{tabular*}{\linewidth}{@{\extracolsep{\fill}}llrr@{}}
\toprule
Turn & Metric & STAIR $-$ Native & STAIR $-$ Vanilla\\
\midrule
\multicolumn{4}{@{}l}{\textbf{\textit{MATH-500}}}\\
T2 & Avg@4 & $+3.10\;[+0.70, +5.75]$ & $-2.40\;[-4.86, +0.03]$\\
T2 & Pass@4 & $+1.20\;[+0.23, +3.31]$ & $+0.80\;[+0.10, +2.56]$\\
T3 & Avg@4 & $+0.90\;[-0.53, +2.54]$ & $+1.45\;[-0.23, +3.57]$\\
T3 & Pass@4 & $+0.20\;[+0.00, +1.08]$ & $+0.80\;[+0.10, +2.61]$\\
T4 & Avg@4 & $+0.45\;[-0.86, +1.80]$ & $+1.75\;[+0.16, +3.79]$\\
T4 & Pass@4 & $+0.00\;[-1.50, +1.36]$ & $+0.40\;[-0.77, +1.87]$\\
Mean & Avg@4 & $+1.48\;[+0.43, +2.68]$ & $+0.27\;[-1.06, +1.79]$\\
Mean & Pass@4 & $+0.47\;[-0.12, +1.34]$ & $+0.67\;[+0.03, +1.78]$\\
\addlinespace
\multicolumn{4}{@{}l}{\textbf{\textit{AIME 2025}}}\\
T2 & Avg@4 & $+12.50\;[-4.05, +29.08]$ & $-18.33\;[-35.17, -4.91]$\\
T2 & Pass@4 & $+23.33\;[+6.14, +50.66]$ & $-3.33\;[-18.81, -0.02]$\\
T3 & Avg@4 & $+5.83\;[-5.34, +19.53]$ & $-10.83\;[-22.99, -0.79]$\\
T3 & Pass@4 & $+13.33\;[+1.72, +37.44]$ & $-3.33\;[-17.43, -0.02]$\\
T4 & Avg@4 & $+16.67\;[+3.07, +32.65]$ & $+3.33\;[-11.61, +19.86]$\\
T4 & Pass@4 & $+3.33\;[-14.67, +24.81]$ & $+0.00\;[-16.14, +15.40]$\\
Mean & Avg@4 & $+11.67\;[+3.16, +20.77]$ & $-8.61\;[-18.72, +1.25]$\\
Mean & Pass@4 & $+13.33\;[+4.00, +26.56]$ & $-2.22\;[-11.97, +3.86]$\\
\addlinespace
\multicolumn{4}{@{}l}{\textbf{\textit{AMC23$^{\dagger}$}}}\\
T2 & Avg@4 & $+8.09\;[-9.74, +26.18]$ & $-16.18\;[-33.34, -0.58]$\\
T2 & Pass@4 & $+8.82\;[-8.66, +31.92]$ & $-2.94\;[-23.06, +11.82]$\\
T3 & Avg@4 & $+11.03\;[+2.90, +21.63]$ & $+8.09\;[-7.09, +28.96]$\\
T3 & Pass@4 & $+0.00\;[+0.00, +0.00]$ & $+11.76\;[+1.45, +34.15]$\\
T4 & Avg@4 & $+0.00\;[-13.45, +14.23]$ & $+2.21\;[-15.74, +20.98]$\\
T4 & Pass@4 & $+0.00\;[-18.51, +19.75]$ & $+5.88\;[-9.77, +27.17]$\\
Mean & Avg@4 & $+6.37\;[-2.18, +15.89]$ & $-1.96\;[-13.86, +11.25]$\\
Mean & Pass@4 & $+2.94\;[-5.86, +14.22]$ & $+4.90\;[-5.24, +18.43]$\\
\addlinespace
\multicolumn{4}{@{}l}{\textbf{\textit{GPQA-Diamond}}}\\
T2 & Avg@4 & $+2.27\;[-2.58, +7.60]$ & $+13.89\;[+7.93, +21.53]$\\
T2 & Pass@4 & $+0.00\;[-6.04, +6.36]$ & $+13.64\;[+6.39, +23.42]$\\
T3 & Avg@4 & $-0.51\;[-5.45, +4.51]$ & $+11.11\;[+4.91, +18.22]$\\
T3 & Pass@4 & $-2.02\;[-9.78, +5.27]$ & $+11.11\;[+3.36, +21.10]$\\
T4 & Avg@4 & $-1.39\;[-5.86, +3.05]$ & $+10.23\;[+4.15, +17.61]$\\
T4 & Pass@4 & $-5.56\;[-12.94, -0.39]$ & $+8.59\;[+1.53, +17.40]$\\
Mean & Avg@4 & $+0.13\;[-2.62, +3.06]$ & $+11.74\;[+7.26, +16.84]$\\
Mean & Pass@4 & $-2.53\;[-6.62, +1.29]$ & $+11.11\;[+5.82, +17.44]$\\
\bottomrule
\end{tabular*}
\end{table}

\clearpage
\begin{table}[H]
\centering
\setlength{\tabcolsep}{8pt}
\renewcommand{\arraystretch}{1.1}
\caption{Qwen3.5-9B: paired differences and pointwise 95\% intervals, in percentage points.}
\label{tab:paired-qwen35_9b_thinking}
\begin{tabular*}{\linewidth}{@{\extracolsep{\fill}}llrr@{}}
\toprule
Turn & Metric & STAIR $-$ Native & STAIR $-$ Vanilla\\
\midrule
\multicolumn{4}{@{}l}{\textbf{\textit{MATH-500}}}\\
T2 & Avg@4 & $+0.85\;[-1.22, +3.03]$ & $-1.50\;[-3.53, +0.54]$\\
T2 & Pass@4 & $+0.00\;[-1.43, +1.32]$ & $+0.60\;[+0.05, +2.14]$\\
T3 & Avg@4 & $-0.20\;[-1.88, +1.46]$ & $+0.45\;[-1.40, +2.62]$\\
T3 & Pass@4 & $-0.20\;[-1.61, +0.84]$ & $+1.20\;[+0.24, +3.42]$\\
T4 & Avg@4 & $+0.55\;[-0.85, +1.93]$ & $+1.45\;[-0.14, +3.63]$\\
T4 & Pass@4 & $+0.00\;[-0.96, +1.05]$ & $+1.00\;[-0.29, +3.27]$\\
Mean & Avg@4 & $+0.40\;[-0.66, +1.46]$ & $+0.13\;[-1.08, +1.62]$\\
Mean & Pass@4 & $-0.07\;[-0.86, +0.67]$ & $+0.93\;[+0.18, +2.30]$\\
\addlinespace
\multicolumn{4}{@{}l}{\textbf{\textit{AIME 2025}}}\\
T2 & Avg@4 & $+4.17\;[-9.51, +18.69]$ & $-10.00\;[-24.92, +2.39]$\\
T2 & Pass@4 & $+3.33\;[-13.26, +24.18]$ & $+0.00\;[-16.18, +16.57]$\\
T3 & Avg@4 & $+8.33\;[-6.58, +24.47]$ & $-14.17\;[-34.91, +2.65]$\\
T3 & Pass@4 & $+3.33\;[-19.24, +28.88]$ & $-3.33\;[-33.41, +26.43]$\\
T4 & Avg@4 & $+5.83\;[-14.46, +24.98]$ & $-12.50\;[-33.50, +7.31]$\\
T4 & Pass@4 & $+6.67\;[+0.27, +26.82]$ & $+3.33\;[-20.88, +29.37]$\\
Mean & Avg@4 & $+6.11\;[-4.16, +16.46]$ & $-12.22\;[-23.71, -0.84]$\\
Mean & Pass@4 & $+4.44\;[-5.67, +17.50]$ & $+0.00\;[-14.89, +17.20]$\\
\addlinespace
\multicolumn{4}{@{}l}{\textbf{\textit{AMC23$^{\dagger}$}}}\\
T2 & Avg@4 & $+4.41\;[-9.26, +16.67]$ & $-6.62\;[-27.73, +11.79]$\\
T2 & Pass@4 & $-5.88\;[-28.36, +10.51]$ & $+0.00\;[-20.21, +19.88]$\\
T3 & Avg@4 & $+7.35\;[-4.99, +19.30]$ & $+2.21\;[-13.02, +22.22]$\\
T3 & Pass@4 & $+2.94\;[-12.03, +21.20]$ & $+8.82\;[+0.68, +28.57]$\\
T4 & Avg@4 & $-2.21\;[-13.26, +8.50]$ & $-2.21\;[-19.89, +18.75]$\\
T4 & Pass@4 & $+5.88\;[+0.21, +23.12]$ & $+11.76\;[+1.54, +34.45]$\\
Mean & Avg@4 & $+3.19\;[-4.91, +10.69]$ & $-2.21\;[-15.10, +11.69]$\\
Mean & Pass@4 & $+0.98\;[-8.34, +10.86]$ & $+6.86\;[-3.14, +21.18]$\\
\addlinespace
\multicolumn{4}{@{}l}{\textbf{\textit{GPQA-Diamond}}}\\
T2 & Avg@4 & $-7.70\;[-13.83, -2.46]$ & $+3.28\;[-2.52, +9.85]$\\
T2 & Pass@4 & $-5.05\;[-13.19, +1.71]$ & $+5.56\;[-2.33, +14.35]$\\
T3 & Avg@4 & $-3.03\;[-7.32, +0.90]$ & $+8.08\;[+2.75, +14.69]$\\
T3 & Pass@4 & $-1.01\;[-7.18, +5.05]$ & $+8.08\;[+1.55, +16.74]$\\
T4 & Avg@4 & $-1.14\;[-6.01, +3.39]$ & $+9.97\;[+4.05, +16.99]$\\
T4 & Pass@4 & $+2.53\;[-2.68, +9.23]$ & $+11.11\;[+4.23, +20.44]$\\
Mean & Avg@4 & $-3.96\;[-6.96, -1.14]$ & $+7.11\;[+3.16, +11.58]$\\
Mean & Pass@4 & $-1.18\;[-4.90, +2.44]$ & $+8.25\;[+3.13, +14.49]$\\
\bottomrule
\end{tabular*}
\end{table}

\clearpage
\subsubsection{Shared-history Controls}\label{app:shared-history-intervals}

For the shared-T1 T2 study, we resample the 30 current problems in 4,000 paired bootstrap replicates. Each sampled problem retains its four responses and, where applicable, all three matched intervention configurations. The same STAIR responses are reused across these configurations. Table~\ref{tab:shared-t1-intervals} reports score differences with pointwise 95\% percentile intervals, conditional on the saved histories and trained controllers.

\begin{table}[H]
\centering
\setlength{\tabcolsep}{7pt}
\renewcommand{\arraystretch}{1.15}
\caption{Qwen3.5-4B on shared-T1 AIME 2025 at T2: paired differences and 95\% intervals (pp). The four-cell interaction follows Equation~\ref{eq:reflector-pairing-interaction} and is measured in Avg@4.}
\label{tab:shared-t1-intervals}
\begin{tabular*}{\linewidth}{@{\extracolsep{\fill}}lrr@{}}
\toprule
Comparison & Avg@4 & Pass@4\\
\midrule
STAIR $-$ Native & $+26.67\;[+16.67,+37.50]$ & $+23.33\;[+10.00,+40.00]$\\
STAIR $-$ Bank-free & $+11.67\;[+2.50,+20.83]$ & $+13.33\;[0.00,+30.00]$\\
STAIR $-$ Random & $+11.11\;[+2.78,+19.72]$ & $+16.67\;[+5.56,+28.89]$\\
STAIR $-$ train/test permutation & $+16.67\;[+8.61,+25.00]$ & $+21.11\;[+7.78,+34.47]$\\
Four-cell interaction & $+15.00\;[+6.39,+24.44]$ & --\\
\bottomrule
\end{tabular*}
\end{table}

Table~\ref{tab:history-management-seeds} reports every decoding-seed group's T4 scores. Table~\ref{tab:history-management} gives their means and sample standard deviations.

\begin{table}[H]
\centering
\setlength{\tabcolsep}{7pt}
\renewcommand{\arraystretch}{1.15}
\caption{Qwen3.5-4B on AIME 2025 at T4 across three decoding-seed groups. Each cell reports Avg@4 / Pass@4 (\%) from 120 responses. All groups use the same Native T1--T3 histories. Bold marks the maximum for each metric within each group.}
\label{tab:history-management-seeds}
\begin{tabular}{lrrr}
\toprule
Condition & Original & Seed 1 & Seed 2\\
\midrule
Native & 49.17 / \textbf{83.33} & 46.67 / 73.33 & 42.50 / 73.33\\
History summarization & 65.83 / 76.67 & \textbf{72.50} / 80.00 & 64.17 / 80.00\\
Turn selection & 35.00 / 66.67 & 38.33 / 70.00 & 35.83 / 60.00\\
Sparse-25\% & 50.83 / 76.67 & 45.83 / 80.00 & 52.50 / 80.00\\
Sparse-50\% & 54.17 / \textbf{83.33} & 52.50 / 70.00 & 46.67 / 76.67\\
Reference-read gate & 59.17 / 76.67 & 55.00 / 80.00 & 57.50 / 86.67\\
\midrule
\rowcolor{stairrow}\textbf{STAIR} & 68.33 / 76.67 & 65.83 / 80.00 & 70.83 / \textbf{90.00}\\
\rowcolor{stairrow}STAIR + Summarization & \textbf{70.00} / 80.00 & 65.00 / 80.00 & 70.83 / 80.00\\
\rowcolor{stairrow}STAIR + Sparse-25\% & 67.50 / 80.00 & 70.83 / 83.33 & \textbf{72.50} / 80.00\\
\rowcolor{stairrow}STAIR + Sparse-50\% & \textbf{70.00} / \textbf{83.33} & \textbf{72.50} / \textbf{86.67} & 66.67 / 76.67\\
\bottomrule
\end{tabular}
\end{table}

\clearpage
\subsubsection{Accuracy Conditional on the First Answer}\label{app:conditional-intervals}

For the pooled conditional accuracies in Table~\ref{tab:history-correctness-summary}, we retain the model--task strata, original response weights, and actual T1 correctness groups. Each of 4,000 crossed multiplier replicates assigns weights to current-question and actual-session clusters. Models sharing a question or session identity receive the same corresponding weight. Table~\ref{tab:history-correctness-intervals} gives the paired conditional-accuracy differences and their 95\% percentile intervals.

\begin{table}[H]
\centering
\setlength{\tabcolsep}{9pt}
\renewcommand{\arraystretch}{1.15}
\caption{Paired conditional-accuracy differences and 95\% intervals (pp), pooled over the three main Qwen models, three mathematical tasks, and T2--T4. $n$ counts responses grouped by actual historical T1 correctness.}
\label{tab:history-correctness-intervals}
\begin{tabular*}{\linewidth}{@{\extracolsep{\fill}}lrrr@{}}
\toprule
Historical T1 & $n$ & STAIR $-$ Native & Native $-$ Vanilla\\
\midrule
Correct & 18,879 & $+0.76\;[+0.15,+1.40]$ & $-0.53\;[-1.48,+0.51]$\\
Wrong & 1,425 & $+8.21\;[+4.34,+12.22]$ & $-19.72\;[-25.08,-14.38]$\\
\bottomrule
\end{tabular*}
\end{table}

\subsubsection{Code-generation Transfer}\label{app:extension-intervals}

For HumanEval+, 4,000 paired resampling replicates preserve question and actual-session dependence and keep each problem's four samples together. Each replicate computes the three later-turn differences before taking their arithmetic mean. Table~\ref{tab:extension-intervals} reports these mean T2--T4 differences and pointwise 95\% percentile intervals.

\begin{table}[H]
\centering
\setlength{\tabcolsep}{9pt}
\renewcommand{\arraystretch}{1.15}
\caption{Qwen3.5-9B on HumanEval+: mean T2--T4 paired differences and 95\% intervals (pp), using all 164 tasks.}
\label{tab:extension-intervals}
\begin{tabular*}{\linewidth}{@{\extracolsep{\fill}}lrr@{}}
\toprule
Comparison & Avg@4 & Pass@4\\
\midrule
STAIR $-$ Native & $+1.73\;[-0.54,+4.23]$ & $+0.41\;[-1.78,+2.91]$\\
STAIR $-$ Vanilla & $+0.41\;[-2.84,+3.90]$ & $+1.63\;[-1.64,+5.76]$\\
\bottomrule
\end{tabular*}
\end{table}

\end{document}